\documentclass{egpubl}
\usepackage{pg2024}

\usepackage[dvipsnames]{xcolor}  %for coloring equations
\usepackage{booktabs} 
\usepackage[linesnumbered,ruled,vlined]{algorithm2e}
\usepackage{lmodern}
\usepackage{anyfontsize}

\makeatletter
\EGlocalpagenumber
\copyrightTextTitPag{\copyright \ 2024}

\def\@oddfoot{{\tiny\raisebox{\z@}[18pt][1pt]{\parbox[b]{20pc}{\sloppy}}}}
\def\@evenfoot{\mbox{}\hfill {\tiny\raisebox{\z@}[8pt][1pt]{\parbox[t]{20pc}{\sloppy}}}}
\makeatother

\CGFStandardLicense
\usepackage[T1]{fontenc}
\usepackage{dfadobe}  
\usepackage{amsmath}
\usepackage{amssymb}
\usepackage{multirow}
\usepackage{float}
\usepackage{cite}  % comment out for biblatex with backend=biber
\BibtexOrBiblatex
\electronicVersion
\PrintedOrElectronic
\ifpdf \usepackage[pdftex]{graphicx} \pdfcompresslevel=9
\else \usepackage[dvips]{graphicx} \fi
\usepackage{egweblnk}

\title[DreamMapping: High-Fidelity Text-to-3D Generation via Variational Distribution Mapping]{DreamMapping: High-Fidelity Text-to-3D Generation via Variational Distribution Mapping}

\author[Cai et al.]
{\parbox{\textwidth}{\centering Zeyu Cai$^{1}$\orcid{0009-0006-5422-4044}
        , Duotun Wang $^{1, 3}$\orcid{0009-0005-4393-5230}
        , Yixun Liang$^{1}$\orcid{0000-0003-4750-8875}
        , Zhijing Shao$^{1}$\orcid{ 0009-0008-3204-3271}
        , Ying-Cong Chen$^{1, 2}$\orcid{0000-0002-9565-8205}
        , Xiaohang Zhan$^{3}$\orcid{0000-0003-2136-7592}
        , and Zeyu Wang$^{1, 2}$\orcid{0000-0001-5374-6330}
        }
        \\
{\parbox{\textwidth}{\centering $^1$The Hong Kong University of Science and Technology (Guangzhou)\\
         $^2$The Hong Kong University of Science and Technology\\
         $^3$Tencent AI Lab\\
       }
}
}
\begin{document}
\teaser{
 \includegraphics[width=0.9\linewidth]{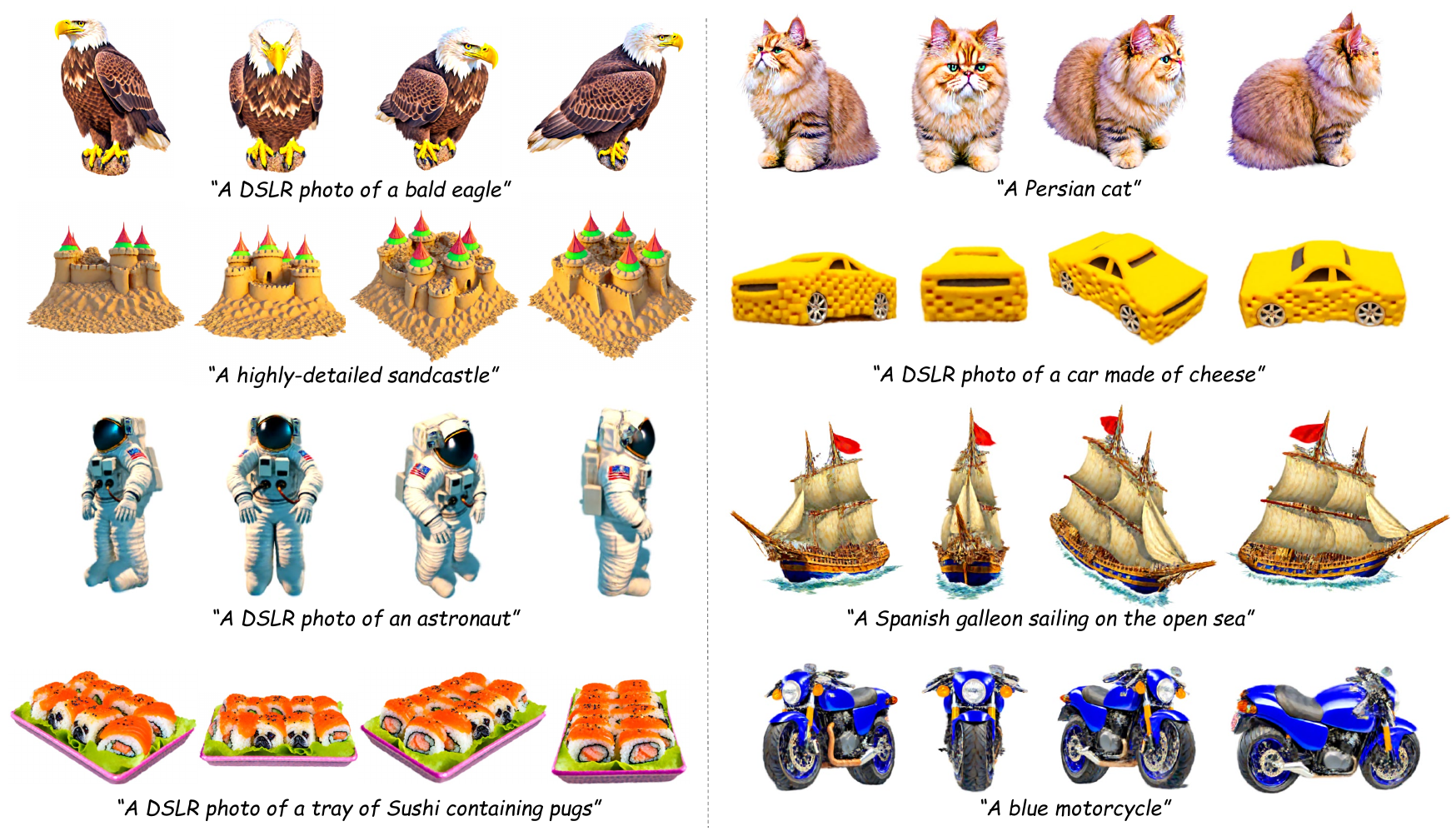}
 \centering
  \caption{Examples of diverse 3D content generated by DreamMapping given text input. Our framework facilitates the rapid distillation of high-fidelity appearance and geometry from pre-trained 2D diffusion models in a short optimization time ($\sim$15 mins on a single A100 GPU).}
\label{fig:teaser}
}

\maketitle
%-------------------------------------------------------------------------
\begin{abstract}
Score Distillation Sampling (SDS) has emerged as a prevalent technique for text-to-3D generation, enabling 3D content creation by distilling view-dependent information from text-to-2D guidance. However, they frequently exhibit shortcomings such as over-saturated color and excess smoothness. In this paper, we conduct a thorough analysis of SDS and refine its formulation, finding that the core design is to model the distribution of rendered images. Following this insight, we introduce a novel strategy called Variational Distribution Mapping (VDM), which expedites the distribution modeling process by regarding the rendered images as instances of degradation from diffusion-based generation. This special design enables the efficient training of variational distribution by skipping the calculations of the Jacobians in the diffusion U-Net. We also introduce timestep-dependent Distribution Coefficient Annealing (DCA) to further improve distilling precision. Leveraging VDM and DCA, we use Gaussian Splatting as the 3D representation and build a text-to-3D generation framework. Extensive experiments and evaluations demonstrate the capability of VDM and DCA to generate high-fidelity and realistic assets with optimization efficiency.
%-------------------------------------------------------------------------
%  ACM CCS 1998
%  (see https://www.acm.org/publications/computing-classification-system/1998)
% \begin{classification} % according to https://www.acm.org/publications/computing-classification-system/1998
% \CCScat{Computer Graphics}{I.3.3}{Picture/Image Generation}{Line and curve generation}
% \end{classification}
%-------------------------------------------------------------------------
%  ACM CCS 2012 (see https://www.acm.org/publications/class-2012)
%The tool at \url{http://dl.acm.org/ccs.cfm} can be used to generate
% CCS codes.
%Example:



\end{abstract}  
\section{Introduction}
There has been an increasing need for digital 3D assets in video games~\cite{VideoGame:book:2012}, mixed reality storytelling~\cite{ARBook:CE:2014, MRBook:ISMAR:2021}, and digital fabrications~\cite{DitalFab:FOA:2012}. However, this widespread growth highlights the need for more efficient 3D content creation, as crafting 3D assets in graphics software is often time-consuming and labor-intensive. Thanks to the rapid development of diffusion models~\cite{DDPM:NIPS:2020, CF_DDPM:NIPS:2021}, recent text-to-3D techniques offer a more accessible and efficient solution for controllable 3D content creation.

Based on the pre-trained text-to-image models~\cite{StableDiffusion:CVPR:2021, DDIM:ICLR:2020}, Score Distillation Sampling (SDS)~\cite{DreamFusion:ICLR:2022, SJC:CVPR:2023} provides pixel-level guidance of 3D assets optimization and has become quite effective and popular for text-to-3D generation. Follow-up works further refine the supervision loss and optimization procedures to produce more realistic and semantic-aligned 3D models~\cite{ProlificDreamer:NIPS:2023, Dreamavatar:Arxiv:2023, Fantasia3D:ICCV:2023, CSD:Arxiv:2023, NFSD:Arxiv:2023, Consistent3D:Arxiv:2024, DreamFlow:Arxiv:2024, LMC-SDS:Arxiv:2024, ASD:Arxiv:2023, Luciddreamer:Arxiv:2023, SSD:Arxiv:2023, LODS:Arxiv:2023}.

% % ?
% However, experimental evidence has revealed that SDS often encounters issues of over-smoothness and color saturation. These challenges significantly hinder the progress of high-fidelity 3D content creation in practical scenarios. Previous studies~\cite{ProlificDreamer:NIPS:2023, Luciddreamer:Arxiv:2023, LMC-SDS:Arxiv:2024, SSD:Arxiv:2023} have probed these problems and attributed them to mode-seeking behaviors in the optimization process of SDS. The core mechanism behind SDS is to align the rendered image, obtained through differentiable rendering, with the image generated by the diffusion model. The intrinsic randomness of the image distribution in the diffusion model inadvertently introduces high variance in the back-propagated guidance gradients, leading to average effects and excess smoothness of generated 3D assets. To mitigate this, SDS employs a substantially large classifier-free guidance (CFG) scale, typically 100. However, this strategy negligently gives rise to color saturation~\cite{ProlificDreamer:NIPS:2023}.\par

In this paper, we review SDS-based 3D generation and its variations, leading to a unifying goal across previous work on refining SDS formulation: establishing a variational distribution for rendered images. Our review and analysis demonstrate that prior efforts have predominantly adopted one of two strategies: 1) utilizing unmodified diffusion models to approximate the variational distribution~\cite{CSD:Arxiv:2023, NFSD:Arxiv:2023, Luciddreamer:Arxiv:2023, Consistent3D:Arxiv:2024} with improved loss designs, and 2) constructing the variational distribution directly through fine-tuning diffusion models~\cite{ProlificDreamer:NIPS:2023, DreamFlow:Arxiv:2024, ASD:Arxiv:2023, LODS:Arxiv:2023} or pre-training a bespoke model~\cite{LMC-SDS:Arxiv:2024}. However, the former approach struggles to represent the distribution accurately, since the rendered images during the optimization process are out-of-domain (OOD) cases as to diffusion models~\cite{NFSD:Arxiv:2023}. Meanwhile, the latter choice demands substantial resources and may exhibit instability during optimization.

\begin{figure}[htb!]
    \centering
    \includegraphics[width=1.0\linewidth]{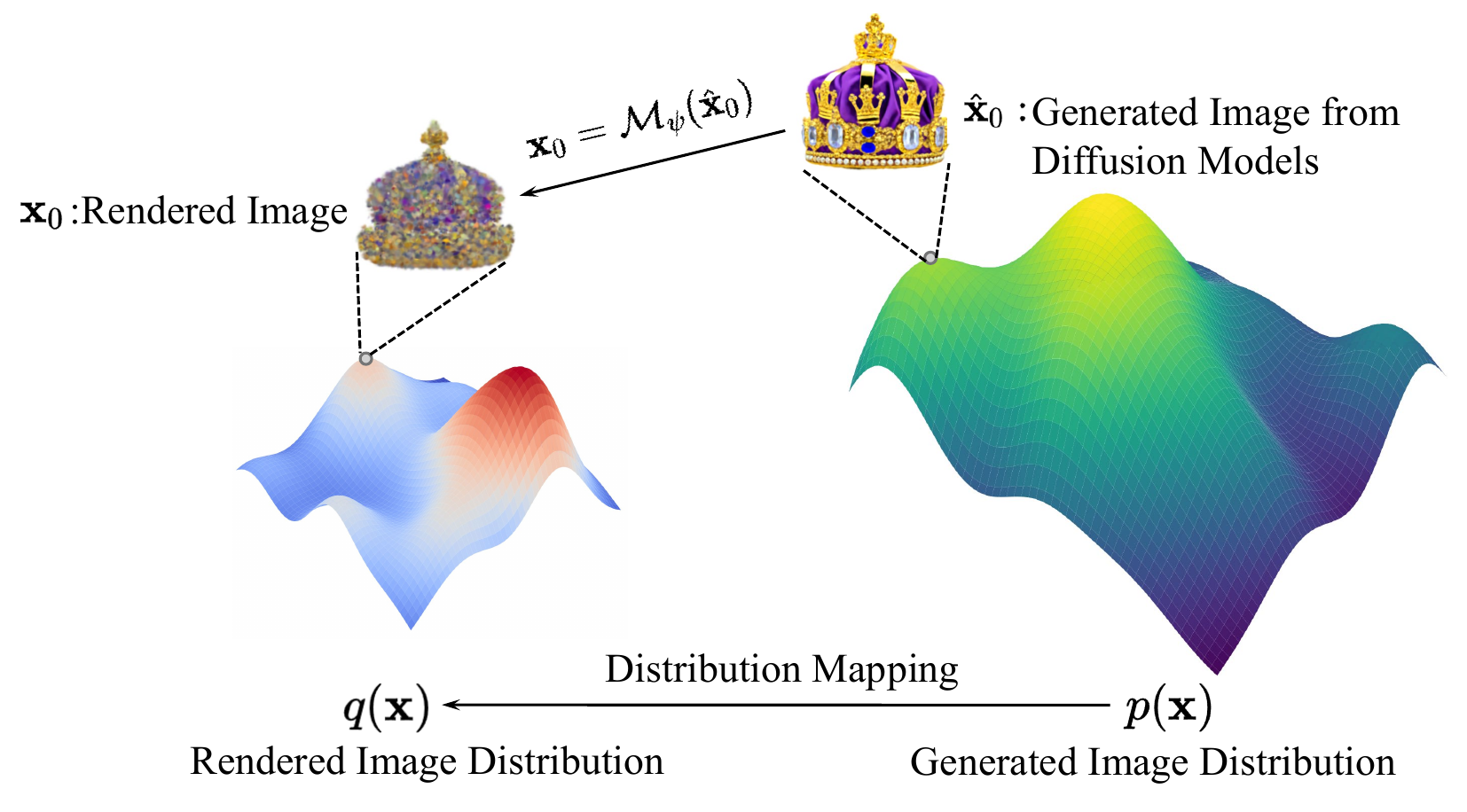}
    \caption{Illustration of our degradation design on the image distributions. We propose the rendered image $\mathbf{x}_0$ can be degraded from the generated image $\mathbf{\hat{x}}_0$ using the trainable degradation operator $\mathcal{M_\psi}$. During SDS optimization, $\mathcal{M}_{\phi}$ can efficiently map the image distributions from $p(\mathbf{x})$ to $q(\mathbf{x})$. Detailed discussions are in Sec.~\ref{sec:vdm}.}
    \label{fig:insights}
\end{figure}

To address these issues, we propose variational distribution mapping (VDM), a novel method treating rendered images as a degraded form of images generated by the diffusion model. VDM efficiently formulates the variational distribution of rendered images by modeling and optimizing the degradation process (i.e., lightweight trainable neural network), which enables us to establish a distribution mapping between the generated and the rendered image distributions (Figure~\ref{fig:insights}). Compared with previous methods, VDM has two notable advantages: 1) By introducing the trainable degradation process, it eliminates the need for the complex Jacobian matrix calculations in the UNet of diffusion model, unlike previous variational distribution modeling approaches~\cite{ASD:Arxiv:2023, LODS:Arxiv:2023}; 2) VDM dynamically models the variational distribution of rendered images, surpassing methods that rely on an unmodified diffusion model with extra batch processing~\cite{NFSD:Arxiv:2023, CSD:Arxiv:2023} for distribution estimation, thereby enhancing generation quality.

% a novel method for efficiently formulating the variational distribution of rendered images. The core mechanism of VDM is based on the key premise that rendered images represent a degraded form of generated images by the diffusion model. This perspective enables us to establish a mapping between the generated and the rendered image distributions by modeling the degradation process. Considering the dynamic optimization of 3D generation and the complexity of degradation, we utilize a trainable neural network to dynamically emulate this process, as depicted in Fig.~\ref{fig:insights}. 

% Our proposed method offers two notable benefits: (1) Introducing the trainable degradation process eliminates the need for the complex Jacobian matrix calculations in the UNet of diffusion model; (2) Our approach to modeling the variational distribution of rendered images dynamically outperforms those using the unmodified diffusion model in accuracy, without requiring additional batch processing (e.g., NFSD~\cite{NFSD:Arxiv:2023, CSD:Arxiv:2023}).

Additionally, we analyze the mode-seeking behavior in SDS and further investigate the probabilistic correlation between the distributions of generated and rendered images. We find that the interdependence of their distributions diminishes as the timestep $t$ decreases. This observation motivates the introduction of the distribution coefficient annealing (DCA) strategy. It applies a time-dependent coefficient to accommodate the dynamic changes of rendered image distribution, thereby improving the generation quality. \par

To demonstrate the generation ability of our proposed methods, we develop a text-to-3D framework incorporating Shap-E~\cite{Shap-e:Arxiv:2023} for initialization and 3D Gaussian Splatting~\cite{3DGS:TOG:2023} as the 3D representation. We present three key contributions as follows:
\begin{itemize}
    \item We systematically review recent SDS-based methods and reveal their limitations, e.g., limited accuracy and efficiency in constructing variational distributions.
    % Recent works that focus on approximating or constructing the variational distributions of rendered images still suffer from accuracy and efficiency issues.
    \item We introduce VDM and the accompanying DCA strategy, which conceptualizes rendered images as degraded diffusion model outputs for rapid variational distribution constructions. Our approach facilitates creating detailed, realistic 3D assets at a reasonable Classifier Free Guidance (CFG) scale~\cite{CFG_DDPM:CVPRW:2021}, i.e., 7.5.
    \item With VDM and DCA, we develop a novel text-to-3D generative framework using Gaussian Splatting, which outperforms existing methods, as demonstrated through extensive evaluations.
\end{itemize}\par

% Importance of 3D assets; With the importance of 3D assets, text-to-3D generation methods are important\par

% Recently, the research on SDS loss and its improvements.\par

% In this paper, we analyze the existing works about SDS and find almost all the works can be concluded as building a variational distribution of rendered images. To solve the mode-seeking problem (VSD)\cite{ProlificDreamer:NIPS:2023}. Based on this insight, we propose ...
\section{Related Work}
\label{sec: related works}
\subsection{Text-to-3D Generation}
\label{sec:related_sds}
With the rapid advancement of diffusion models and 3D representations, significant progress has been made in text-based generation. DreamFields~\cite{DreamFields:CVPR:2022} introduced the optimization of Neural Radiance Fields (NeRF) using guidance from pre-trained CLIP models, but the quality was limited due to the CLIP loss's insufficient semantic guidance. Building on this, DreamFusion~\cite{DreamFusion:ICLR:2022} and SJC~\cite{SJC:CVPR:2023} developed SDS, a novel loss function based on probability density distillation that provides pixel-level guidance by seeking specific modes in a text-guide diffusion model, enhancing the quality and efficiency of optimization-based 3D generation.\par

The advent of SDS marked a turning point that inspired a series of subsequent works in 3D generation~\cite{ProlificDreamer:NIPS:2023, CSD:Arxiv:2023, NFSD:Arxiv:2023, DreamFlow:Arxiv:2024, Luciddreamer:Arxiv:2023, LODS:Arxiv:2023, ASD:Arxiv:2023}. These studies have refined the generation of 3D assets from various perspectives. Some utilizes advanced differentiable 3D representations to enhance outcomes~\cite{Magic3D:CVPR:2023, Fantasia3D:ICCV:2023}, while others~\cite{Consistent3D:Arxiv:2024}focus on mitigating the ``Janus'' problem~\cite{Debias:NIPS:2023, LMC-SDS:Arxiv:2024}. Zero-1-to-3~\cite{Zero123:ICCV:2023} and MVDream~\cite{MVDream:Arxiv:2023} fine-tuned the diffusion model with multi-view image datasets to improve 3D consistency. Perp-Neg~\cite{Perp-Neg:Arxiv:2023} alleviates the Janus problem with view-dependent negative prompts, and ESD~\cite{ESD:Arxiv:2023} introduces a view-conditioned loss to improve multi-view generation.\par

A central emphasis in recent research lies in refining the original SDS methodology, aiming to secure more precise guidance from diffusion models. ProlificDreamer~\cite{ProlificDreamer:NIPS:2023} introduces a Variational Score Distillation Sampling technique to alleviate the mode collapse issue inherent to SDS. To further mitigate over-saturation and over-smoothness problems, TextMesh~\cite{Textmesh:3DV:2024} enforces textured optimizations with high SDS gradients only and integrates multi-view consistent diffusion. Make-it-3D~\cite{Make3D:2023:ICCV} proposes two-stage optimizations to enhance textured appearance; Fantasia3D~\cite{Fantasia3D:ICCV:2023} dynamically adjusts the time-dependent weighting function in SDS computations. Meanwhile, CSD~\cite{CSD:Arxiv:2023} and NFSD~\cite{NFSD:Arxiv:2023} demonstrate the pivotal role of the conditional term in SDS for 3D generation. They employ additional negative prompts to refine the optimization process. Their explorations highly inspired our work, and we propose further improving SDS with the trainable construction of variational distribution. We will continue more in-depth discussions in Sec.~\ref{sec:revisit}. 

\subsection{Differentiable 3D Representations}
The differentiable 3D representations are pivotal to text-to-3D generation. A 3D model parameterized by $\theta$ is rendered into an image from camera viewpoint $c$ using the differentiable rendering function $g(\theta, c)$. This allows for optimizing the 3D models through backpropagation to align with pixel-level guidance derived from diffusion models. Numerous differentiable 3D representations have been utilized in prior text-to-3D works~\cite{DreamFusion:ICLR:2022, Magic3D:CVPR:2023, Latent-NeRF:CVPR:2022, SIGGRAPH:TextDeformer:2023, DreamGaussian:Arxiv:2023}, including NeRF~\cite{NeRF:ECCV:2020} which is frequently employed in generation tasks. However, NeRF's volume rendering is computationally intensive, posing challenges for rendering high-resolution images efficiently enough to leverage the diffusion model's guidance.\par

To mitigate this constraint, DMTet~\cite{DMTet:NIPS:2021} has been introduced to integrate both explicit and implicit representations, which has gained increasing utilizations~\cite{Fantasia3D:ICCV:2023, HeadArtist:Arxiv:2023}. Additionally, there is a growing interest in applying purely explicit representations to facilitate smooth shape manipulations within graphics software. This involves a gradient-based mesh deformation approach to create 3D models~\cite{SIGGRAPH:TextDeformer:2023, HeadEvolver:Arxiv:2024}. Notably, 3D Gaussian Splatting~\cite{3DGS:TOG:2023} has emerged as an efficient and high-quality explicit representation for reconstruction tasks and has been incorporated into several text-to-3D generation works~\cite{DreamGaussian:Arxiv:2023, GaussianDreamer:Arxiv:2023, Luciddreamer:Arxiv:2023}. We investigate Gaussian Splatting as the 3D representation for our generation framework.

% \subsection{Diffusion Models}
% Diffusion models have revolutionized the realm of image synthesis and manipulation with their superior generative capabilities. The influence of Denoising Diffusion Probabilistic Model (DDPM) is particularly noteworthy, as they adeptly generate and refine imagery based on textual cues. These models iteratively corrupt images with noise in a controlled manner, akin to a Markov chain, and are then trained to reverse this process, effectively restoring the original image. Amongst these models, Stable Diffusion represents a leading-edge development, utilizing latent diffusion techniques to boost both the efficiency of training and the sampling process. Its prowess in text-guided image manipulation has proven to be a cornerstone for the latest advancements in text-to-3D generation technologies.

\section{Preliminaries}
In this section, we briefly introduce diffusion models and SDS, which are the fundamental theories of this work.

\subsection{Diffusion Models}
The Diffusion Model~\cite{DDPM:NIPS:2020,DDIM:ICLR:2020,Score-Based-Model:ICLR:2021} is a likelihood-based generative model designed to approximate the data distribution $p_{data}$ starting from gaussian noise. Given a sample $\mathbf{x}$ from $p_{data}$, the model undergoes a forward diffusion process over a series of timesteps $t\in[0, t_{max}]$, whereby noise is incrementally added to transform $\mathbf{x}$ into Gaussian noise $\mathbf{x}_{T} \sim \mathcal{N}(0,I)$. The diffusion at each step is described by:
\begin{equation}
    p(\mathbf{x}_{t}|\mathbf{x}_{t-1}) = \mathcal{N}(\mathbf{x}_{t};\sqrt{1-\beta_{t}}\mathbf{x}_{t-1}, \beta_{t}I),
\end{equation}
where $\beta_{t}$ is a predetermined noise schedule and $\mathbf{x}_{0} = \mathbf{x} \sim p_{data}$. To reverse this process, a learnable UNet~\cite{UNet:MICCAI:2015} with parameters $\phi$ estimates the posterior:
\begin{equation}
    p_{\phi}(\mathbf{x}_{t-1}|\mathbf{x}_t) = \mathcal{N}(\mathbf{x}_{t-1}; \sqrt{\bar{\alpha}_{t-1}}\mu_{\phi}(\mathbf{x}_{t}), (1-\bar{\alpha}_{t-1})\Sigma_{\phi}(\mathbf{x}_{t})),
\end{equation}
with $\mu_{\phi}(\mathbf{x}_{t})$ and $\Sigma_{\phi}(\mathbf{x}_{t})$ representing the mean and variance predictions for $\mathbf{x}_{t}$, and $\bar{\alpha}_{t}:=\prod_{1}^{t}(1-\beta_{t})$.

The training goal is to optimize $\mu_{\phi}(\mathbf{x}_{t})$ and $\Sigma_{\phi}(\mathbf{x}_{t})$ to maximize the log likelihood's variational lower bound. In practice, the learning target is re-parameterized to the added noise $\epsilon \sim \mathcal{N}(0, I)$ which is used to produce $\mathbf{x}_{t}$ from $\mathbf{x} = \mathbf{x}_{0}$, and the notation of UNet in diffusion models is $\epsilon_{\phi}(\mathbf{x}_{t}, t)$. The learning process of diffusion models can be interpreted as predicting the noise $\epsilon$ that corrupts the data $\mathbf{x}$. Furthermore, recent studies~\cite{Score-Based-Model:ICLR:2021, NCSN:NIPS:2019} have shown that $\epsilon_{\phi}(\mathbf{x}_{t}, t)$ corresponds to the score function $\nabla_{\mathbf{x}_{t}}\log p_{\phi}(\mathbf{x}_{t})$, indicating the model's ability to guide $\mathbf{x}_{t}$ towards regions of higher density within $p_{\phi}(\mathbf{x_{t}})$, which can be approximated as:
\begin{equation}
    \nabla_{\mathbf{x}_{t}}\log p_{\phi}(\mathbf{x}_{t}) \approx -\frac{1}{\sqrt{1-\bar{\alpha}_{t}}}\epsilon_{\phi}(\mathbf{x}_{t}, t).
    \label{Score Relationship}
\end{equation}

\textbf{Classifier Free Guidance:} Compared with unconditional image generation, text-guided image generation~\cite{CFG_DDPM:CVPRW:2021, Imagen:NIPS:2022, StableDiffusion:CVPR:2021} has a higher demand. Conditioned on the text prompt $y$, diffusion models accept it as another input for the diffusion process, denoted as $\epsilon_{\phi}(\mathbf{x}_{t},t,y)$, with related score function $\nabla_{\mathbf{x}_{t}}\log p(\mathbf{x}_{t}|y)$. Classifier Free Guidance (CFG)~\cite{CFG_DDPM:CVPRW:2021} is utilized as an implicit classifier to get the textual guidance for image generation. It has a changeable hyperparameter named CFG scale, hereafter denoted as $s$, and the original prediction is changed as:
\begin{equation}
    \epsilon_{\phi}^{\text{CFG}}(\mathbf{x}_{t},t,y) = \epsilon_{\phi}(\mathbf{x}_{t},t,y) + s\cdot (\epsilon_{\phi}(\mathbf{x}_{t},t,y) - \epsilon_{\phi}(\mathbf{x}_{t},t)).
    \label{SD: CFG}
\end{equation}
As mentioned in Eq. (\ref{Score Relationship}), this prediction is also related to a corresponding score function using Bayes' rule:
\begin{equation}
    \epsilon_{\phi}^{\text{CFG}}(\mathbf{x}_{t},t,y) \propto \nabla_{\mathbf{x}_{t}}\log p_{\phi}(\mathbf{x}_{t}|y) + s \cdot\nabla_{\mathbf{x}_{t}}\log p_{\phi}(y|\mathbf{x}_{t}).
    \label{Eq:CFG:Score}
\end{equation}
the latter term can be intuitively understood as guiding $x_{t}$ being more in line with the text description $y$.

\subsection{SDS with Differentiable Rendering}
\label{pre:sds}
As demonstrated in Sec.~\ref{sec:related_sds}, SDS~\cite{DreamFusion:ICLR:2022, SJC:CVPR:2023}  is an optimization-driven 3D generative technique that leverages pre-trained 2D diffusion models. Let $\mathbf{x}_{0}=g(\theta,c)$ represent the rendered image derived from differentiable 3D representations, parameterized by $\theta$ and camera pose $c$. The distribution of noisy rendered images is then formulated as follows:
\begin{equation}
    \label{eq:x_t define}
    q^{\theta}(\mathbf{x}_{t}) = \mathcal{N}(\mathbf{x}_{t};\sqrt{\bar{\alpha}_{t}}\mathbf{x}_{0},(1-\bar{\alpha}_{t})I).
\end{equation}

SDS adopts text-conditioned noisy real image distribution $p_{\phi}(\mathbf{x}_{t}|y)$ represented by pre-trained Stable Diffusion~\cite{StableDiffusion:CVPR:2021} and optimizes the parameter $\theta$ by minimizing the following KL divergence for all timestep $t$:
\begin{equation}
    \label{SDS:KL}
    \min_{\theta \in \Theta}\mathcal{L}_{SDS}(\theta) := \mathbb{E}_{t,c} \left[ \omega(t)D_{KL}(q^{\theta}(\mathbf{x}_{t})||p_{\phi}(\mathbf{x}_{t}|y)) \right],
\end{equation}
where $\omega(t)$ is a time-dependent weighting function. Eq. (\ref{SDS:KL}) can be further rewritten as the gradient of a weighted probability density distillation loss to update $\theta$:
\begin{equation}
\label{sds:grad}
    \nabla_{\theta}\mathcal{L}_{SDS}(\theta) \approx \mathbb{E}_{t,\epsilon,c}[ \omega(t)(\epsilon_{\phi}(\mathbf{x}_{t},t,y)-\epsilon)\frac{\partial \mathbf{x}}{\partial \theta}],
\end{equation}

where $\epsilon \sim \mathcal{N}(0,I)$ and the UNet Jacobian term $\frac{\partial \epsilon_{\phi}(\mathbf{x}_{t},t,y)}{\partial \mathbf{x}_{t}}$ is ignored based on the analysis in DreamFusion~\cite{DreamFusion:ICLR:2022}.

% systematic review table
\begin{table*}[!tbh]
\footnotesize
\centering
\caption{Summarize and categorize advancements in SDS. Recent studies primarily concentrate on enhancing SDS through unconditional term refinement. Developing a more adaptable and efficient variational distribution for rendered images improves generation quality.}

\begin{tabular}{ ccc}
\hline
\multicolumn{1}{c}{\textbf{Method}} & \multicolumn{1}{c}{\textbf{Unconditional Term}} & \multicolumn{1}{c}{\textbf{Core Mechanism}}                                                                                     \\ \hline
SDS~\cite{DreamFusion:ICLR:2022}          &    $\epsilon_{\phi}(\mathbf{x}_{t},t,y)-\epsilon$                & Distillation baseline                                                                                         \\ \hline
VSD~\cite{ProlificDreamer:NIPS:2023}          &  $\epsilon_{\phi}(\mathbf{x}_{t},t,y)-\epsilon_{\text{lora}}(\mathbf{x}_{t},t,y)$                  & \multirow{5}{*}{\begin{tabular}[c]{@{}c@{}}Construct the variational distribution \\ directly with extra models\end{tabular}}                                                               \\ %\cline{1-2}
DreamFlow~\cite{DreamFlow:Arxiv:2024}    &    $\epsilon_{\phi}(\mathbf{x}_{t},t,y)-\epsilon_{\text{lora}}(\mathbf{x}_{t},t,y)$                  &                                                                                                  \\ %\cline{1-2}
ASD~\cite{ASD:Arxiv:2023}          &    $\epsilon_{\phi}(\mathbf{x}_{t},t,y)-\epsilon_{\text{lora}}(\mathbf{x}_{t},t,y)$                  &                                                                                                  \\ %\cline{1-2}
LODS~\cite{LODS:Arxiv:2023}         &     $\epsilon_{\phi}(\mathbf{x}_{t},t,y)-\epsilon_{\text{lora}}(\mathbf{x}_{t},t,y)$  or  $\epsilon_{\phi}(\mathbf{x}_{t},t,y)-\epsilon_{\phi}(\mathbf{x}_{t},t,y_{\psi})$                 &                                                                                                  \\ %\cline{1-2}
LMC-SDS~\cite{LMC-SDS:Arxiv:2024}      &  $\epsilon_{\phi}(\mathbf{x}_{t},t,y)-b_{\psi}(\mathbf{x}_{t},t)$      &                                                                                                  \\ \hline
CSD~\cite{CSD:Arxiv:2023}          &   $\epsilon_{\phi}(\mathbf{x}_{t},t,y)-\lambda_{s}\epsilon_{\phi}(\mathbf{x}_{t},t,y_{neg})$      & \multirow{5}{*}{\begin{tabular}[c]{@{}c@{}}Approximate the variational distribution \\ from original Stable Diffusion\end{tabular}} \\ %\cline{1-2}
NFSD~\cite{NFSD:Arxiv:2023}         &      $\epsilon_{\phi}(\mathbf{x}_{t},t,y)-\lambda_{t}\epsilon_{\phi}(\mathbf{x}_{t},t,y_{neg})$              &                                                                                                  \\ %\cline{1-2}
ISM~\cite{Luciddreamer:Arxiv:2023}          &    $\epsilon_{\phi}(\mathbf{x}_{t},t,y)-\epsilon_{\phi}(\mathbf{x}_{s},s)$                &                                                                                                  \\ %\cline{1-2}
SSD~\cite{SSD:Arxiv:2023}          &  $\lambda_{t}(\epsilon_{\phi}(\mathbf{x}_{t},t,y) - r(\mathbf{x}_{t},\epsilon,t,y)\cdot\epsilon) $            &                                                                                                  \\ %\cline{1-2}
Consistent3D~\cite{Consistent3D:Arxiv:2024} &    $\epsilon_{\phi}(\mathbf{x}_{t_1},t_1,y)-\epsilon_{\phi}(\mathbf{x}_{t_2},t_{2},y)$                &                                                                                                  \\ \hline
\end{tabular}
\label{tab:survey}
\end{table*}
\section{Methodology}
\subsection{Systematic Review of Advancements in SDS}
\label{sec:revisit}
% Notes: Separate conditional and unconditional term, since the conditional term has a reasonable score function, related works are all focus on the unconditional term.

As outlined in Sec.~\ref{pre:sds}, SDS employs a formal gradient, detailed in Eq. (\ref{sds:grad}), to optimize the 3D representation during generation. To achieve text-guided 3D generation, SDS integrates textual information through the common CFG form, as shown in Eq. (\ref{SD: CFG}). Consequently, the practical SDS gradient expression is:
\begin{equation}
\label{sds:grad:CFG}
    \nabla_{\theta}\mathcal{L}_{SDS}(\theta) = \mathbb{E}_{t,\epsilon,c}[\omega(t)(\epsilon_{\phi}^{CFG}(\mathbf{x}_{t},t,y)-\epsilon)\frac{\partial \mathbf{x}}{\partial \theta}],
\end{equation}
and the noise residual can be further formed as:
\begin{multline}
\label{cond:uncond}
      \epsilon_{\phi}^{CFG}(\mathbf{x}_{t},t,y)-\epsilon = \underbrace{\epsilon_{\phi}(\mathbf{x}_{t},t,y) - \epsilon }_{\text{unconditional term}}+ \\
  s \cdot \underbrace{(\epsilon_{\phi}(\mathbf{x}_{t},t,y) - \epsilon_{\phi}(\mathbf{x}_{t},t))}_{\text{conditional term}},
\end{multline}
% determined by an implicit classifier 
which delineates two components: the unconditional and the conditional terms. Numerous studies have indicated that the conditional term pivotal in ensuring the rendered image aligns with textual semantics~\cite{ProlificDreamer:NIPS:2023, CSD:Arxiv:2023, NFSD:Arxiv:2023, SSD:Arxiv:2023}. This term is further guided by the score function $\nabla_{\mathbf{x}_{t}}\log p_{\phi}(y|\mathbf{x}_{t})$, as indicated in Eq. (\ref{Eq:CFG:Score}). Consequently, subsequent research mostly focuses on refining the unconditional term to improve SDS.

%Many research efforts (e.g., ) have been refining the unconditional term since it provides the necessary direction for the image to conform to the text semantics.

The unconditional term, known for its propensity to introduce over-smoothing artifacts due to mode-seeking characteristics, can be isolated from Eq. (\ref{sds:grad:CFG}) as follows:
\begin{equation}
\label{sds:grad:uncond}
    \nabla_{\theta}\mathcal{L}_{SDS}^{uncond}(\theta) = \mathbb{E}_{t,\epsilon,c}[\omega(t)(\epsilon_{\phi}(\mathbf{x}_{t},t,y)-\epsilon)\frac{\partial \mathbf{x}}{\partial \theta}].
\end{equation}
Given that $\epsilon$ represents zero-mean Gaussian noise, the expected value of the product $\mathbb{E}_{t,\epsilon,c}[\omega(t)(-\epsilon)]\frac{\partial \mathbf{x}}{\partial \theta}$ is zero. Therefore, the gradient for updating the 3D parameters $\theta$ simplifies to $\mathbb{E}_{t,\epsilon,c}[\omega(t)\epsilon_{\phi}(\mathbf{x}_{t},t,y)\frac{\partial \mathbf{x}}{\partial \theta}]$, which steers $\theta$ towards the modes of the conditional posterior $p_{\phi}(\mathbf{x}_{t}|y)$. However, as SDS involves a multi-step optimization process, the mode achieved is an average of samples drawn from $p_{\phi}(\mathbf{x}_{t}|y)$ for all timestep $t$, resulting in over-smoothed generative outcomes~\cite{ProlificDreamer:NIPS:2023}.

Numerous studies have proposed various strategies to address the challenge of mode-seeking in 3D generation. Our review of recent advancements in refining the standard SDS approach reveals that these efforts generally focus on creating a variational distribution for the rendered images, which shifts the mode-seeking operation to minimizing the KL divergence between two distinct distributions. As illustrated in Table \ref{tab:survey}, VSD~\cite{ProlificDreamer:NIPS:2023} uses text-conditioned particles (or rendered images when the number of particles equals one) to build a variational distribution of 3D representations. VSD employs a LoRA~\cite{LoRA:ICLR:2021} model to swiftly optimize the score of rendered images, termed as $\epsilon_{\text{lora}}(\mathbf{x}_{t},t,y)$. DreamFlow~\cite{DreamFlow:Arxiv:2024} follows VSD with a similar design, guided by theories of Schrodinger Bridge~\cite{I2SB:ICML:2023}. ASD~\cite{ASD:Arxiv:2023} enhances the optimization methods of VSD within the framework of GAN~\cite{GAN:NIPS:2014}, while LODS~\cite{LODS:Arxiv:2023} diversifies the modeling methods through learnable text embeddings denoted as $y_{\psi}$. LMC-SDS~\cite{LMC-SDS:Arxiv:2024} pre-trains a neural network to learn the score manifold corrective transitioning rendered images to generated images. The aforementioned studies collectively endeavor to determine the variational distribution of rendered images directly. Despite employing parameter-efficient modeling techniques, i.e., LoRA, the optimization of variational distribution requires additional computation of the UNet Jacobian for Stable Diffusion, which is time-consuming and prone to instability. For instance, given a text prompt, VSD, ASD, and LODs require approximately 8 hours, 5 hours, and 2 hours, respectively to complete the 3D generation. In contrast, DreamFusion only necessitates 1.5 hours since it skips the additional computations on the diffusion model.

Another line of research uses the original Stable Diffusion to approximate the rendered image distribution. CSD~\cite{CSD:Arxiv:2023} and NFSD~\cite{NFSD:Arxiv:2023} obtain an empirical solution by using negative prompt conditioned posterior of Stable Diffusion to represent the rendered image distribution. To accommodate the dynamic characteristics of generation, they use iteration steps related weights $\lambda_{s}$ and timesteps related weights $\lambda_{t}$ separately. ISM~\cite{Luciddreamer:Arxiv:2023} uses DDIM inversion~\cite{DDIM:ICLR:2020} to get the score of rendered images. SSD~\cite{SSD:Arxiv:2023} applies a closed-form solution $r(\mathbf{x}_{t},\epsilon,t,y)$ to project 3D reprentations to the generated image domain of diffusion models. In Consistent3D~\cite{Consistent3D:Arxiv:2024}, the score of Stable Diffusion in smaller timestep $t_{2}$ is treated as the rendered image distribution score. Although this line of research approximates rendered image distribution in a training-free way, the final results may not always be satisfying. Since rendered images during the optimization process are commonly changeable and unnatural, being out-of-domain for the generated image distribution modeled by diffusion models, taking a precise negative prompt or applying a closed-form modeling strategy is unable to model the rendered image generation correctly. Worse still, an additional inference at each optimization iteration is introduced due to negative prompts (e.g., CSD~\cite{CSD:Arxiv:2023} and NFSD~\cite{NFSD:Arxiv:2023}).

\subsection{Variational Distribution Mapping}
\label{sec:vdm}

During the iterative optimization of SDS, the inferior optimization of 3D object parameters often compromises the quality of rendered images, leading to low-quality (LQ) renderings. Diffusion models face challenges in directly representing low-quality (LQ) images due to limited training in such cases. However, considering the potent capabilities of diffusion models in generating high-quality (HQ) images, it's reasonable to assume they can approximate the HQ counterparts of rendered LQ images by treating the latter as degraded versions of the former. Specifically, given $\hat{\mathbf{x}}_{0} \sim p(x)$ as a sample of diffusion output, the rendered image $\mathbf{x}_{0}=g(\theta,c)$ can be represented by a degradation result:
\begin{equation}
    \mathbf{x}_{0} = \mathcal{M} (\hat{\mathbf{x}}_{0}) + n, 
    \label{eq:degradation process orig}
\end{equation}
where $ \mathcal{M}$ is the degradation operator and $n$ is the observation noise. Taking diffusion model outputs as real images, the process in Eq. (\ref{eq:degradation process orig}) is a kind of typical real image degradation process~\cite{DPS:ICLR:2022}.

\begin{figure}[htb!]
    \centering
    \includegraphics[width=1.0\linewidth]{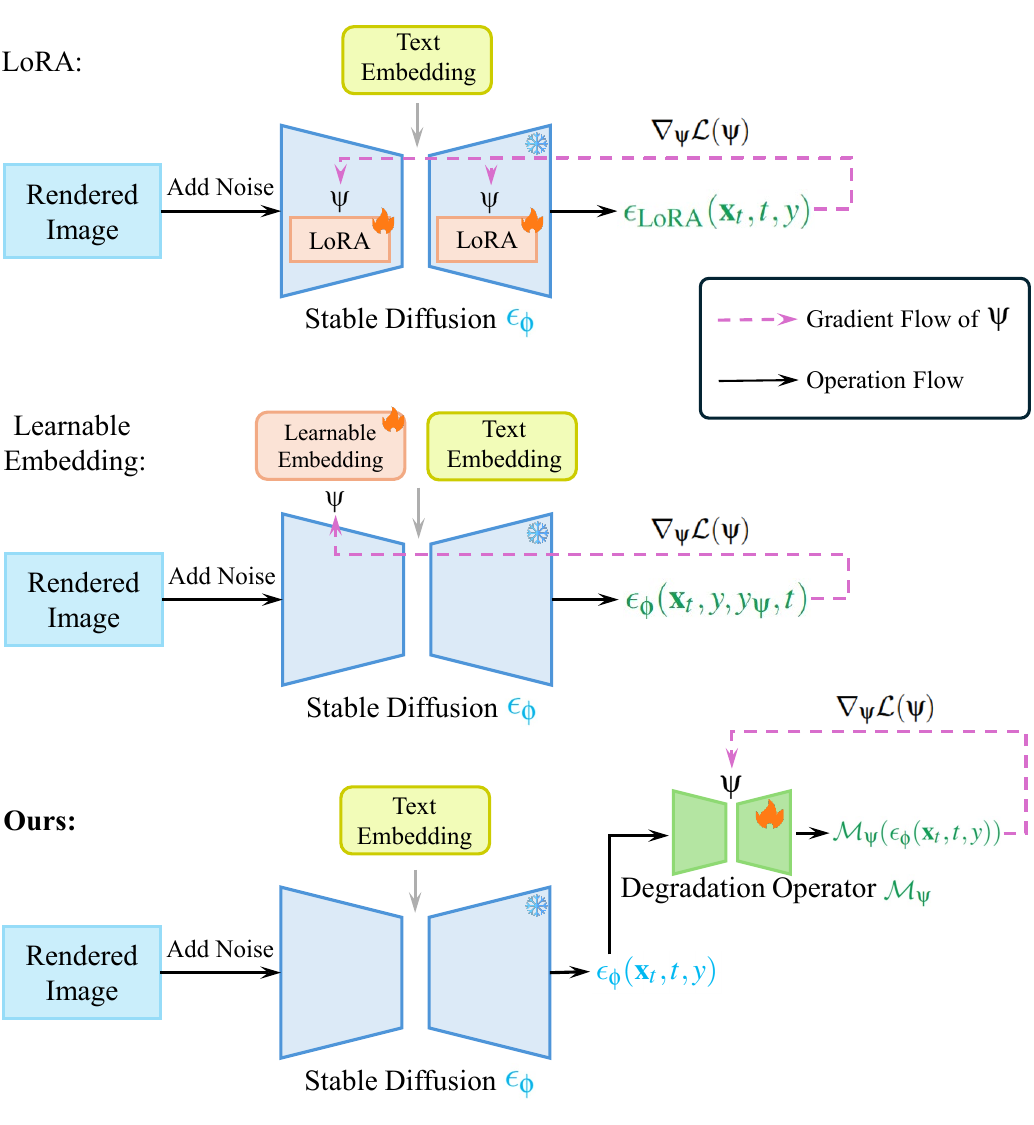}
    \caption{Advantage of VDM. Other solutions modeling the variational distribution of rendered images require extra time to calculate the complex UNet Jacobian matrix in diffusion models. For instance, methods applying LoRA~\cite{ProlificDreamer:NIPS:2023, DreamFlow:Arxiv:2024, ASD:Arxiv:2023, LODS:Arxiv:2023} and Learnable Embedding~\cite{LODS:Arxiv:2023}, when optimizing the variational distribution, the gradient backward must pass through the Stable Diffusion UNet, leading to extra computing time. Our VDM overcomes this problem, taking less time to optimize.}
    \label{fig:VDM difference}
\end{figure}

\begin{figure*}[!htb]
    \centering
    \includegraphics[width=0.95\linewidth]{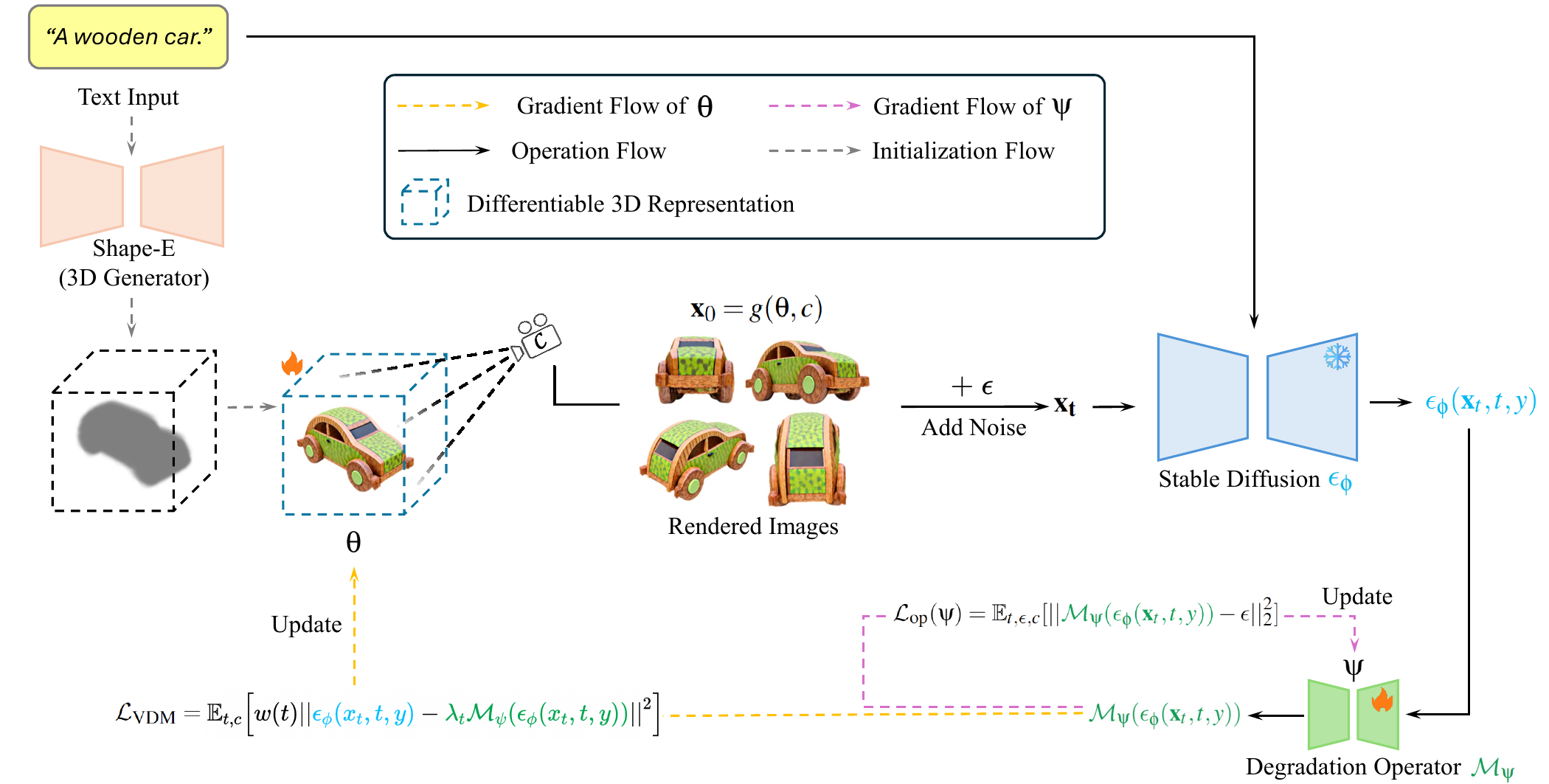}
    \caption{Framework overview. In our text-to-3D generation, we start with the shape initialization (i.e., Shape-E~\cite{UnpairedShape:Graph:2018}) of the 3D representations $\theta$ based on the text input $y$. By incorporating pre-trained Stable Diffusion, we disturb rendered images of random views $\mathbf{x}=g(\theta,c)$ to noisy latents $\mathbf{x}_t$. After learning the image degradation $\psi$, we update $\theta$ with the VDM-based loss $\mathcal{L}_{VDM}$. It is worth noting that the gradient flows bypass the frozen UNet Jacobian terms of Stable Diffusion, significantly expediting the optimization process.}
    % Please refer to Secs.~\ref{sec:vdm} and ~\ref{sec:dca} for technical details.
    \label{fig:overview}
\end{figure*}

However, SDS-based 3D generation is a multi-step optimization process with different timestep $t$, a singular fixed degradation operator cannot represent all the situations. Therefore, as shown in Figure~\ref{fig:insights}, we apply a neural network $\mathcal{M}_\psi$ to model this complex degradation process by learning network weights $\psi$. According to the experimental results, we omit observation noise $n$ and use $\mathcal{M}_{\psi}$ to represent the non-linear degradation operator. The learnable degradation process is further formed as:
\begin{equation}
    \label{eq:degradation process}
    \mathbf{x}_{0} = \mathcal{M}_{\psi}(\hat{\mathbf{x}}_{0}),
\end{equation}
where $\hat{\mathbf{x}}_{0}$ is a sample of the diffusion model generation results, with the posterior $p_{\phi}(\mathbf{x}_{0}|y)$ given the text prompt $y$. The degraded image $\mathbf{x}_{0}$ belongs to a learnable variational distribution $q_{\psi}(\mathbf{x}_{0}|\hat{\mathbf{x}}_{0})$. Our goal is to minimize the KL divergence of rendered images and degraded image distribution, with the formulation as:
\begin{equation}
    \label{eq:KL op X0}
    \min_{\psi}D_{KL}(q^{\theta}(\mathbf{x_{0}})||q_{\psi}(\mathbf{x}_{0}|\hat{\mathbf{x}}_{0})).
\end{equation}
Following ProlificDreamer~\cite{ProlificDreamer:NIPS:2023}, Eq. (\ref{eq:KL op X0}) can be transferred into a series of optimization problems with different diffused distributions indexed by timestep $t$. For an arbitrary $t$, $\psi$ will be optimized from:
\begin{equation}
    \label{eq:KL op Xt}
    \min_{\psi}D_{KL}(q^{\theta}(\mathbf{x_{t}})||q_{\psi}(\mathbf{x}_{t}|\hat{\mathbf{x}}_{t})),
\end{equation}
where $\mathbf{x}_{t} = \sqrt{\bar{\alpha}}\mathbf{x}_{0} + \sqrt{1 - \bar{\alpha}}\epsilon$ and $\hat{\mathbf{x}}_{t} = \sqrt{\bar{\alpha}}\mathbf{x}_{0} + \sqrt{1 - \bar{\alpha}}\epsilon_{\phi}(\mathbf{x}_{t}, t, y)$. Similar to Eq. (\ref{SDS:KL}) to Eq. (\ref{sds:grad}), the gradient loss  $\nabla_{\psi}\mathcal{L}_{\text{op}}$ to optimize the degradation operator $\mathcal{M}_{\psi}$ is: 
\begin{equation}
    \label{eq:loss op x}
    \nabla_{\psi}\mathcal{L}_{\text{op}}(\psi) := \nabla_{\psi}\mathbb{E}_{t,\epsilon,c}[||\mathcal{M}_{\psi}(\hat{\mathbf{x}}_{t}) - \mathbf{x}_{t}||_{2}^{2}].
\end{equation}
Since $\bar{\alpha}$ is a constant pre-defined in diffusion models, for easy calculation, Eq. (\ref{eq:loss op x}) is equivalent to noise correction~\cite{DDPM:NIPS:2020}:
\begin{equation}
    \label{eq:loss op epsilon}
    \nabla_{\psi}\mathcal{L}_{\text{op}}(\psi) = \nabla_{\psi}\mathbb{E}_{t,\epsilon,c}[||\mathcal{M}_{\psi}(\epsilon_{\phi}(\textbf{x}_{t},t,y)) - \epsilon||_{2}^{2}]. 
\end{equation}

Through optimizing the degradation operator, we learn a variational distribution mapping from the distribution of diffusion models prediction to the rendered image distribution. Compared to other variational distribution modeling methods, a key advantage of VDM is that the optimization of $\mathcal{M}_{\psi}$ does not require complex Jacobian matrix calculations in the UNet of the diffusion model, as illustrated in Figure~\ref{fig:VDM difference}.

\subsection{Distribution Coefficient Annealing}
\label{sec:dca}
\begin{figure}[!htb]
    \centering
    \includegraphics[width=0.95\linewidth]{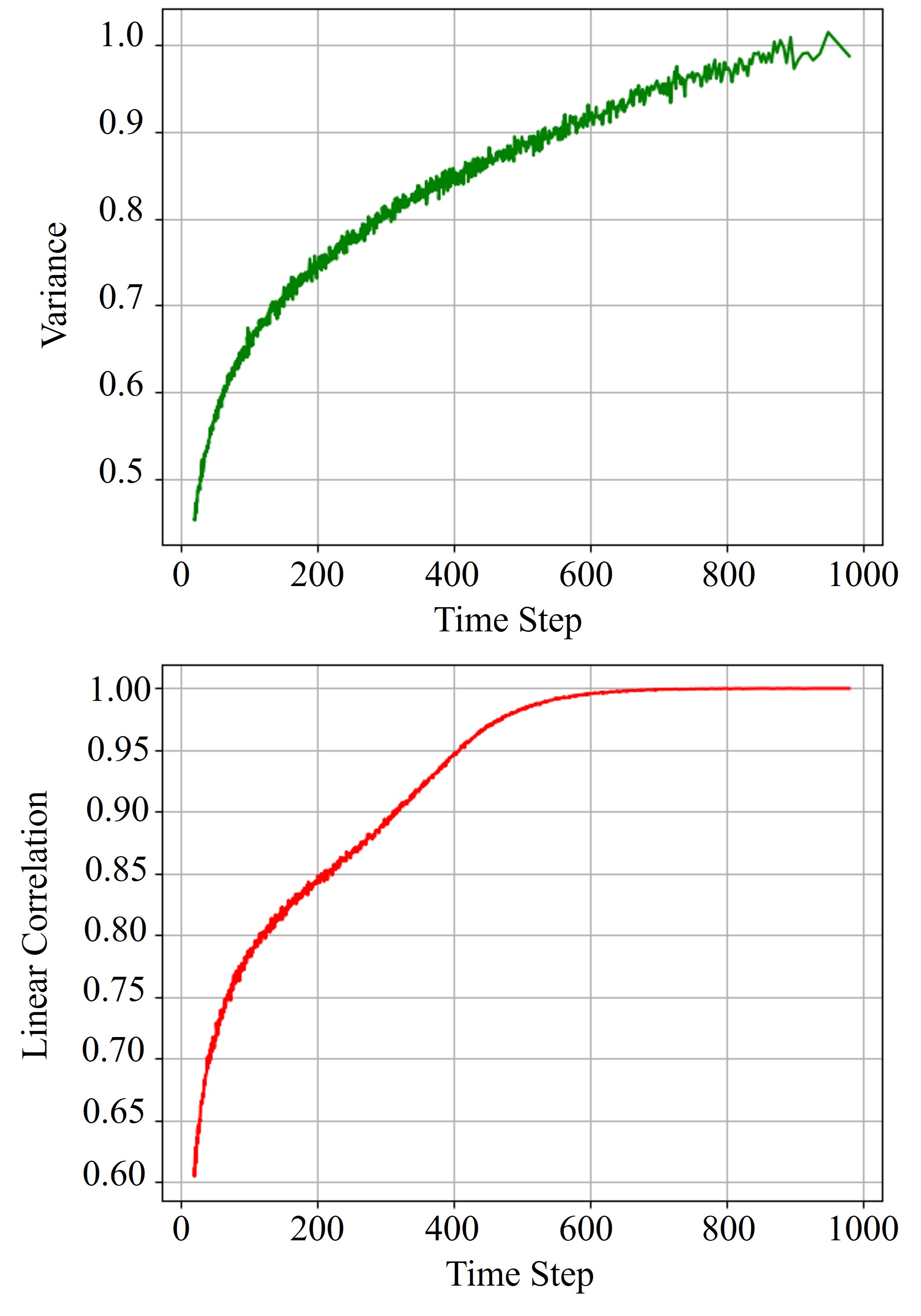}
    \caption{Variance of $\epsilon_{\phi}(\mathbf{x}_{t}, t, y)$ and linear correlation between $\epsilon$ and $\epsilon_{\phi}(\textbf{x}_{t},t,y)$ over time step $t$.}
    \label{fig:variance of noise prediction}
\end{figure}

We further investigate into $\epsilon_{\phi}(\textbf{x}_{t},t,y)$ with its corresponding score function  $\nabla_{\mathbf{x}_{t}}\log p(\mathbf{x}_{t}|y)$, analyzing its properties and finding different features in different timesteps. Specifically, when the timestep becomes large, $t \rightarrow T \text{and }\bar{\alpha}_{t} \rightarrow 0$. The added noise $\sqrt{1-\bar{\alpha}_{t}}\epsilon$ tends to have a relatively large value and high variance. Based on this, if we seek a special mode from the posterior $p_{\phi}(\mathbf{x}_{t}|y)$ to guide the optimization of 3D objects at large timesteps, the guidance will also have high variance and lead to over-smoothing results (refer to Figure~\ref{fig:variance of noise prediction}). 
% As analyzed in Sec.~\ref{sec:revisit}, SDS mitigates this artifact by taking an abnormally large value of the CFG scale (i.e., 100), and other solutions~\cite{SSD:Arxiv:2023, LODS:Arxiv:2023} take rendered images as a variational distribution. 
When the timestep $t$ is small, however, the mode-seeking behavior is precisely what we want to retrieve since $p_{\phi}(\mathbf{x}_{t}|y) \approx p_{\phi}(\mathbf{x}|y)$.\par

Additionally, we find that our design of VDM may not work perfectly at small timesteps, as the modeling of the degradation process builds on the assumption that at least $\epsilon$ and $\epsilon_{\phi}$ are associated. Nonetheless, the linear correlation between $\epsilon$ and $\epsilon_{\phi}(\textbf{x}_{t},t,y)$ as defined in~\cite{Colinear:IS:2022} is anticipated to diminish as $t$ decreases (shown in Figure~\ref{fig:variance of noise prediction} and proved in Sec.~\ref{sec: analysis for DCA}), which means $\epsilon_{\phi}(\textbf{x}_{t},t,y)$ becomes independent of $\epsilon$ when $t$ is small~\cite{SSD:Arxiv:2023}. Consequently, VDM may not flawlessly work alone in small timesteps, where mode-seeking is beneficial, as well as $\epsilon_{\phi}(\textbf{x}_{t},t,y)$ and $\epsilon$ are unrelated. To address this, we propose the distribution coefficient annealing (DCA) strategy for better 3D generation, which adds a time-dependent coefficient $\lambda_{t}$ to the rendered image distribution. $\lambda_{t}$ is calculated by:
\begin{equation}
    \label{eq:DCA}
        \lambda_{t} = \begin{cases}
        1, & t>300 \\
        1-\bar{\alpha}_{t}, & t \leq 300,
    \end{cases}
\end{equation}
and the final gradient of VDM loss over 3D parameters $\theta$ is:
\begin{equation}
    \footnotesize
    \nabla_{\theta}\mathcal{L}_{\text{VDM}}= \mathbb{E}_{t,\epsilon,c}[\omega(t)(\epsilon_{\phi}(\textbf{x}_{t},t,y) - \lambda_{t}\mathcal{M}_{\psi}(\epsilon_{\phi}(\textbf{x}_{t},t,y)))\frac{\partial \mathbf{x}_{t}}{\partial \theta}].
\end{equation}

The Jacobian term $\frac{\partial \epsilon_{\phi}(\textbf{x}_{t},t,y) - \lambda_{t}\mathcal{M}_{\psi}(\epsilon_{\phi}(\textbf{x}_{t},t,y)}{\partial \mathbf{x}_{t}}$ is also ignored following the conclusion of SDS~\cite{DreamFusion:ICLR:2022}. 

\subsection{Generation Framework}
Following recent text-to-3D generation methods~\cite{DreamFusion:ICLR:2022, ProlificDreamer:NIPS:2023, Fantasia3D:ICCV:2023}, our optimized-based framework leverages 3D Gaussian Splatting (3DGS), which is highly efficient in rendering to high-fidelity 3D creations. As Liang et al.~\cite{Luciddreamer:Arxiv:2023} have demonstrated, point cloud initialization is critical for geometry quality. We adopt Shape-E~\cite{Shap-e:Arxiv:2023} to generate the coarse initialization with shape prior (Figure~\ref{fig:overview}). Algorithm \ref{ag:pipe} shows the 3D generation process of DreamMapping.

\begin{algorithm}[!htb]
\caption{The generation pipeline of DreamMapping}
\KwIn{Large text-to-image diffusion model $\epsilon_{\phi}$, 3D object initialized by Shap-E with parameters $\theta$, learnable operator $M_{\psi}$, text prompt $y$}
\label{ag:pipe}
\BlankLine 
\While{\textnormal{$\theta$ not converged}}{
  \emph{Randomly simple a camera pose $c$.}\;
  \emph{Render the 3D object at pose $c$ to get a 2D image $\mathbf{x}_{0}=g(\theta, c)$}\;
  \emph{Sample $\epsilon \sim \mathcal{N}(0, I), t \sim \mathcal{U}(0, 1000)$}\;
  \emph{$\mathbf{x}_{t} = \sqrt{\bar{\alpha}}_{t}\mathbf{x} + \sqrt{1-\bar{\alpha}_{t}}\epsilon$}\;
  \If{t < 300}{
    \emph{$\lambda_{t} = 1-\bar{\alpha}_{t}$}\;
  }
  \Else{
    \emph{$\lambda_{t} = 1$}\;
  }
  \emph{$\mathcal{L}_{\text{op}} = \mathbb{E}_{t,\epsilon,c}[||\mathcal{M}_{\psi}(\epsilon_{\phi}(\textbf{x}_{t},t,y)) - \epsilon||_{2}^{2}] $}\;
  \emph{$\mathcal{L}_{\text{VDM}}= \mathbb{E}_{t,\epsilon,c}[\omega(t)||\epsilon_{\phi}(\textbf{x}_{t},t,y) - \lambda_{t}\mathcal{M}_{\psi}(\epsilon_{\phi}(\textbf{x}_{t},t,y))||_{2}^{2}] $}\;
  \emph{Update $\psi$ with $\nabla_{\psi}\mathcal{L}_{\text{op}}$}\;
  \emph{Update $\theta$ with $\nabla_{\theta}\mathcal{L}_{\text{VDM}}$}\;

  % \ifthenelse{}
}   
\end{algorithm}

%\begin{equation}
%    \mathcal{L}_{\text{op}}(\psi) = \mathbb{E}_{t,\epsilon,c}[||\textcolor{ForestGreen}{\mathcal{M}_{\psi}(\epsilon_{\phi}(\textbf{x}_{t},t,y))} - \epsilon||_{2}^{2}] 
%\end{equation}

%\begin{equation}
%        \nabla_{\theta}\mathcal{L}_{\text{VDM}}= \mathbb{E}_{t,\epsilon,c}[\omega(t)||\textcolor{Cyan}{\epsilon_{\phi}(\textbf{x}_{t},t,y)} - \textcolor{ForestGreen}{\lambda_t \mathcal{M}_{\psi}(\epsilon_{\phi}(\textbf{x}_{t},t,y))}||_{2}^{2}\frac{\partial \textbf{x}_{t}}{\partial \theta}] 
%\end{equation}

\section{Experiments}
\begin{table*}[hbt!]
\centering
\caption{Quantitative comparison with text-to-3D methods on generation consistency. The CLIP score~\cite{CLIP:ICML:2021} measures the semantic similarity between text prompts and randomly rendered views. Generation time measures averaged time cost per text prompt. ViT-L/14 and ViT-bigG-14 represent two different backbones used to calculate the CLIP score.}
\begin{tabular}{cccc}
\hline
Method        & ViT-L/14 ($\uparrow$) & ViT-bigG-14 ($\uparrow$) & Generation Time($\downarrow$) \\ \hline
Shap-E~\cite{Shap-e:Arxiv:2023} & 20.51       & 32.21         & 6 seconds     \\
DreamFusion~\cite{DreamFusion:ICLR:2022}     & 23.60       & 37.46          & 1.5 hours       \\
DreamGaussian~\cite{DreamGaussian:Arxiv:2023}       & 20.85      & 30.51         & 5 minutes      \\
GaussianDreamer~\cite{GaussianDreamer:Arxiv:2023} & 27.23       & 41.88          & 15 minutes      \\
ProlificDreamer~\cite{ProlificDreamer:NIPS:2023} & \underline{27.39}       & \textbf{42.98}          & 10 hours        \\
Ours            & \textbf{30.16}       & \underline{42.10}        & 15 minutes      \\ \hline
\end{tabular}
\label{tab:clip}
\end{table*}

\begin{figure*}[h!]
\centering\includegraphics[width=1.0\linewidth]{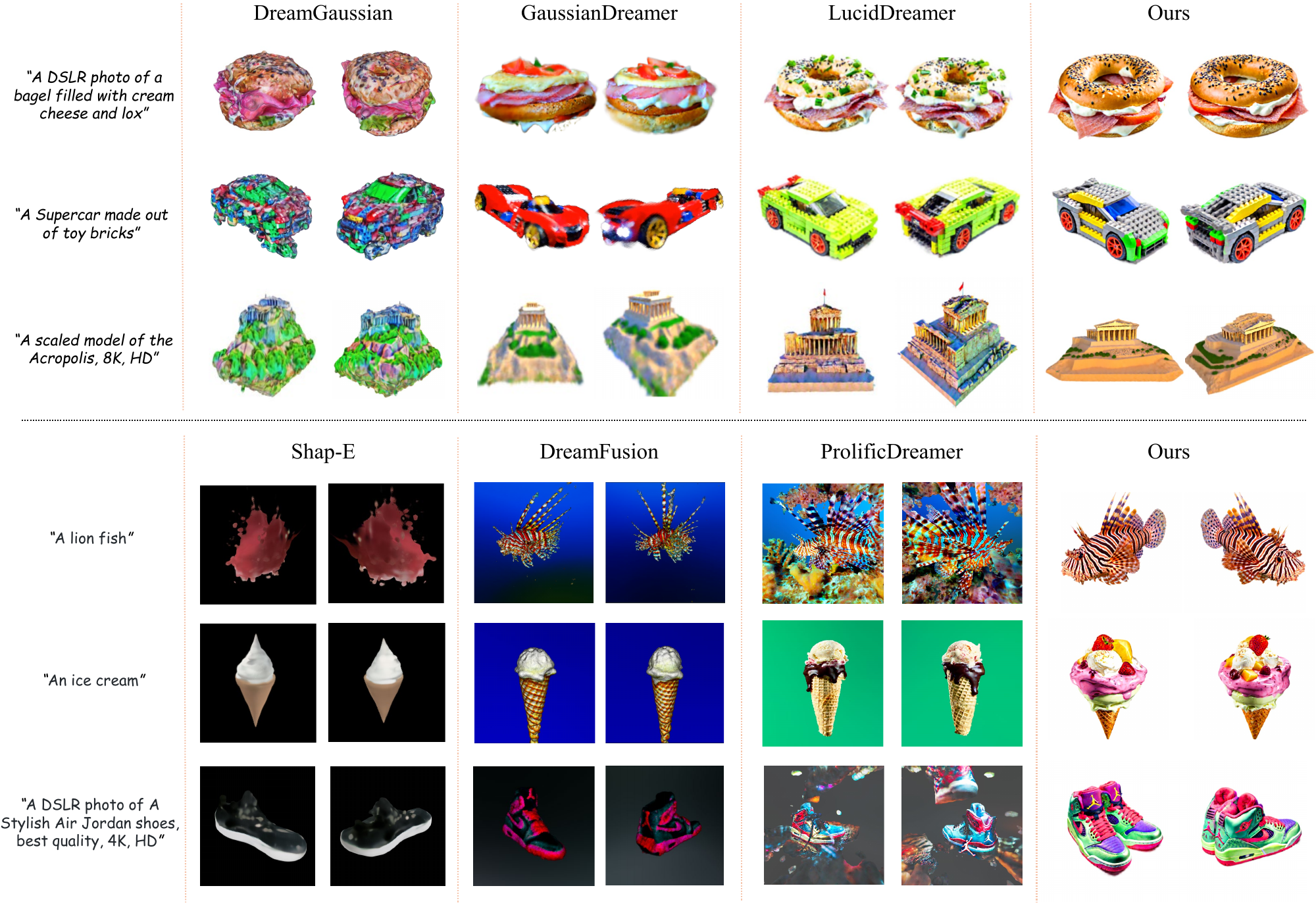}
    \caption{Qualitative comparisons with recent popular methods in text-to-3D generation based on 3DGS and NeRF. We present rendered images of two views for each method. Experimental results demonstrate that our method generates 3D content closely aligned with textual prompts, exhibiting high fidelity and intricate details. Please zoom in for details. Additional comparisons can be found in Figure~\ref{fig:add_vis_comp}.}
    \label{fig:qualitative}
\end{figure*}

The diverse 3D models generated through our framework are demonstrated in Figures~\ref{fig:teaser} and~\ref{fig:add_res}. Our framework can produce 3D assets that align closely with textual semantics, presenting fine appearances and detailed features. It effectively handles creative long-text descriptions, such as ``A Spanish galleon sailing on the open sea'' (Figure~\ref{fig:teaser}), while mitigating excess smoothness and color saturation issues, for instance, in fur textures or layered burritos. This section first discusses the choice of 3D representations and outlines the implementation details of our generation framework, followed by experiments and a user study to compare our outcomes with those produced by state-of-the-art (SoTA) methods. An ablation study is then conducted to assess the effectiveness of key design choices in our approach.

%color changed 
\subsection{Choice of 3D Representation}
In this study, we used 3DGS as the primary 3D representation and also showed the efficacy of our proposed method on NeRF (Figure~\ref{fig:ablation NeRF}). Compared to NeRF, 3DGS offers several advantages:
\begin{itemize}
    \item Efficiency. 3DGS outperforms NeRF in rendering and optimization efficiency (i.e., less generation time and lower GPU usage). ProlificDreamer uses NeRF for high-quality 3D generation but is time-consuming (10 hours), as shown in Table~\ref{tab:clip}.
    \item Explicity. 3DGS is an explicit 3D representation and can be adapted to traditional graphics pipelines. Recent works have successfully applied 3DGS in Unity~\cite{VRGS:SIGGRAPH:2024, Splattingavatar:CVPR:2024}.
    \item Versatility. Several concurrent works (e.g., 2DGS~\cite{2DGS:SIGGRAPHl:2024} and GoF~\cite{GOF:arXiv:2024}) have been put forward to improve the geometry structure of 3DGS. These studies demonstrate optimization and rendering efficiency comparable to the original 3DGS. However, improvements to NeRF such as Mip-NeRF360~\cite{MipNeRF:CVPR:2022} present an unacceptable long training time.
\end{itemize}

\subsection{Implemention Details}
We employed the codebase from LucidDreamer ~\cite{Luciddreamer:Arxiv:2023} and replaced the ISM loss with our proposed VDM and DCA. Stable-Diffusion-2-1-Base (SD 2.1 Base)~\cite{SD-2-1-Base} was utilized as the base diffusion model $\epsilon_{\phi}$. For the geometry initialization, we used Shap-E~\cite{Shap-e:Arxiv:2023} as the prior point cloud generator and upsampling the number of initialization points to nearly 50000. The batch size is set to 4, and the optimization iterations for each 3D object is 5000. For the learnable operator $\mathcal{M}_{\psi}$, we selected U-Lite~\cite{ULite-APSIPA-2023}, a kind of UNet~\cite{UNet:MICCAI:2015} with only $1M$ parameters in the learnable neural network. The learning rate of $\mathcal{M}_{\psi}$ is $0.01$ to efficiently model the degradation process.

\subsection{Quantitative Results}
We performed comprehensive experiments to assess the semantic coherence (CLIP-score) and visual quality (3D-FID) of the generated 3D content. The quantitative results presented in Table~\ref{tab:clip} and Table~\ref{tab:fid} indicate that our method outperforms state-of-the-art 3DGS-based methods in terms of rendering appearance and multi-view semantic consistency. Although ProlificDreamer attains the highest CLIP score on ViT-bigG-14~\cite{CLIP:ICML:2021}, its generation time is considerably longer than DreamFusion, which also employs the NeRF representation. This is attributed to the time-consuming process of constructing a variational distribution in ProlificDreamer. In contrast, our approach efficiently constructs a variational distribution with a generation time comparable to GaussianDreamer, which does not build such a distribution. To further demonstrate the efficiency and quality of our methods, we implemented NeRF with VDM and DCA (seen in Figure~\ref{fig:ablation NeRF}), achieving a generation time of approximately 1.9 hours per text prompt on a single A100 GPU. We believe our quantitative evaluation will inform future research utilizing 3DGS-based generation.

\begin{table}[htb!]
\caption{Quantitative comparison with text-to-3D methods on generation appearance. The 3D-FID~\cite{ProlificDreamer:NIPS:2023} is evaluated between rendered images of random views and reference images sampled by 50-step DPM-Solver++~\cite{DPMSolver:Arxiv:2023}. $\downarrow$ indicates lower is better.}
\footnotesize   % incase not overflow
\centering
\begin{tabular}{c c}
         \hline 
              Method & 3D-FID $\downarrow$ (25 Prompts)\\
         \hline 
         SDS~\cite{DreamFusion:ICLR:2022} &  $ 191.82$ \\
         ProlificDreamer~\cite{ProlificDreamer:NIPS:2023} & $159.84$ \\
         DreamGaussian~\cite{DreamGaussian:Arxiv:2023} & $265.88$  \\
         GaussianDreamer~\cite{GaussianDreamer:Arxiv:2023} & $156.71$  \\
         LucidDreamer~\cite{Luciddreamer:Arxiv:2023} & \underline{142.63}  \\
         Ours & \textbf{139.03} \\
         \hline
    \end{tabular}
\label{tab:fid}
\end{table}

\begin{table}[thb]\centering
    \caption{User preferences. Small values indicate high rankings. Our results achieved the highest ranking.} 
    \label{tab:userstudy}
    \resizebox{0.48\textwidth}{!}{
    \large
    \begin{tabular}{*{10}{c}}
        \toprule
        & Semantic Alignment &  Generation Quality\\
        \midrule
        DreamGaussian~\cite{DreamGaussian:Arxiv:2023}  & 4.94 & 4.83 \\
        ProlificDreamer~\cite{ProlificDreamer:NIPS:2023} & \underline{2.50} & \underline{2.34}\\ 
        GaussianDreamer~\cite{GaussianDreamer:Arxiv:2023}  & 3.72  & 3.80 \\
         LucidDreamer~\cite{Luciddreamer:Arxiv:2023} & 2.78 & 2.54 \\
         Ours  & \textbf{1.05}  & \textbf{1.49} \\
        %\multicolumn{5}{c}{Semantic Alignment} \\
        \bottomrule
    \end{tabular}
    }
\end{table}

\subsection{Qualitative Results}
We compared our model with current SoTA baselines using 3DGS representation (i.e., DreamGaussian~\cite{DreamGaussian:Arxiv:2023}, GaussianDreamer~\cite{GaussianDreamer:Arxiv:2023}, LucidDreamer~\cite{Luciddreamer:Arxiv:2023}). SD 2.1 Base was used for distillation, and all generation experiments were conducted on Nvidia A100 for fair comparison. As shown in Figure~\ref{fig:qualitative}, we can see that DreasmGaussian suffers from low quality with blurry appearance and incomplete shape. Results of GaussianDreamer and LucidDreamer show good semantic consistency and realistic colors. However, they still generate vague textures on parts of generated models. For example, the tail light part in the case of ``A supercar made out of toy bricks'' is blurry, and fur details are missing in GaussianDreamer. Instead, our approach efficiently predicts an accurate variational distribution of rendered images, yielding diverse semantically meaningful 3D assets with clear geometry structure and intricate appearance details. Lastly, it is important to highlight that LucidDreamer's generation time is typically longer than our approach due to its use of the DDIM inversion method~\cite{DDIM:ICLR:2020}, which requires substantially more computational resources, such as loss computations. Please refer to Figure~\ref{fig:add_res} more visual results.  

\subsection{User Study}
To present real-world user preferences, we conducted a user study to provide a comprehensive evaluation. Specifically, we select 28 text prompts from ViT-L/14~\cite{CLIP:ICML:2021} to render multiple views using different text-to-3D frameworks. 36 Users, of whom 26 are graduate students majoring in computer graphics and vision and 10 are company employees specializing in AI content generation, were invited to rank them based on the rendering fidelity and the degree of semantic alignment with given text descriptions. As shown in Table~\ref{tab:userstudy}, our framework achieves the highest average ranking, indicating that our approach significantly outperforms existing methods in text-to-3D tasks with respect to human preferences.
%refer to dreamfusion

\begin{figure}[htb!]
    \centering
    \includegraphics[width=1.0\linewidth]{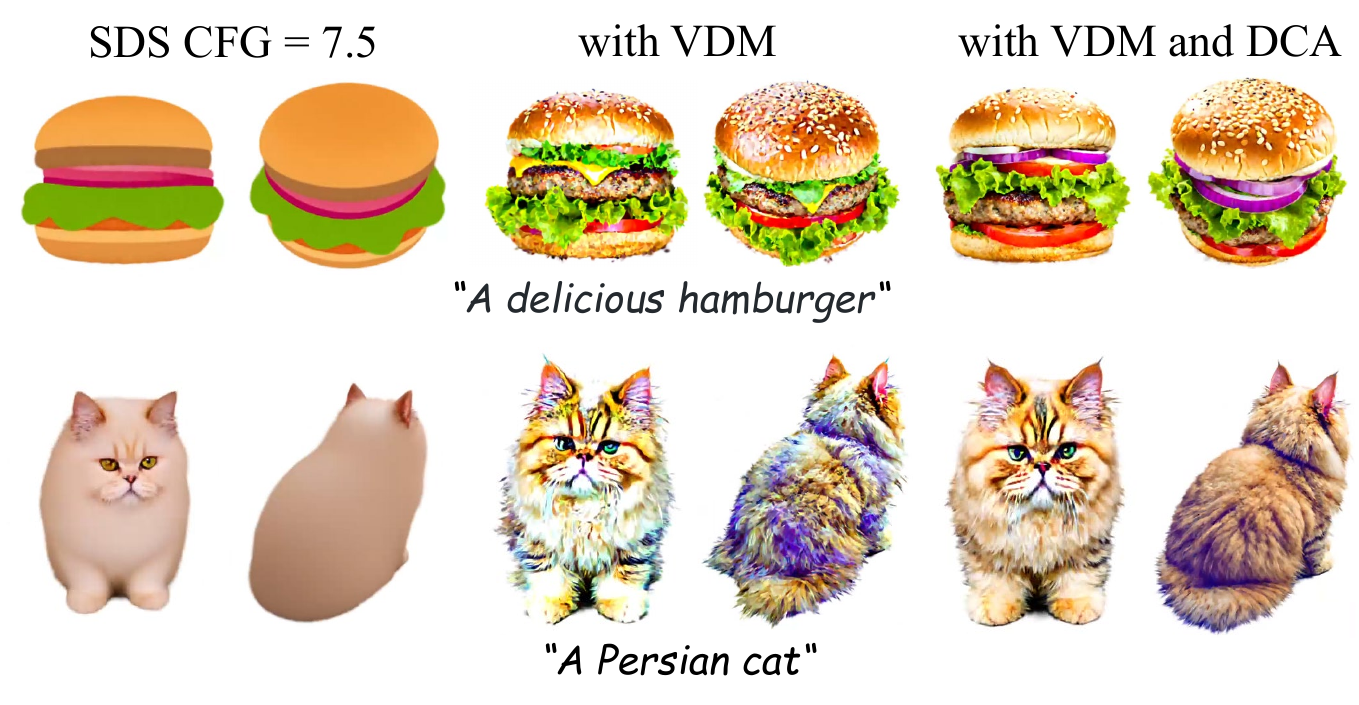}
    \caption{Ablation study on EDM and DCA. Compared to SDS, EDM significantly adds appearance details to 3D models, and DCA further controls color saturation.}
    \label{fig:ablation_1}
\end{figure}

\subsection{Ablation Study}
\label{sec:ablation}
\textbf{Effect of VDM and DCA.} 
We examine the impact of our proposed VDM and DCA. As illustrated in Figure~\ref{fig:ablation_1}, when the CFG scale is set to $7.5$, the original SDS produces overly smooth 3D appearance. Incorporating VDM enhances the visual details of generated 3D assets but leads to slight color saturation and persistent noise issues. By integrating both VDM and DCA, our framework is capable of generating highly detailed and realistic 3D models.\par

\begin{figure}[htb!]
    \centering
    \includegraphics[width=1.0\linewidth]{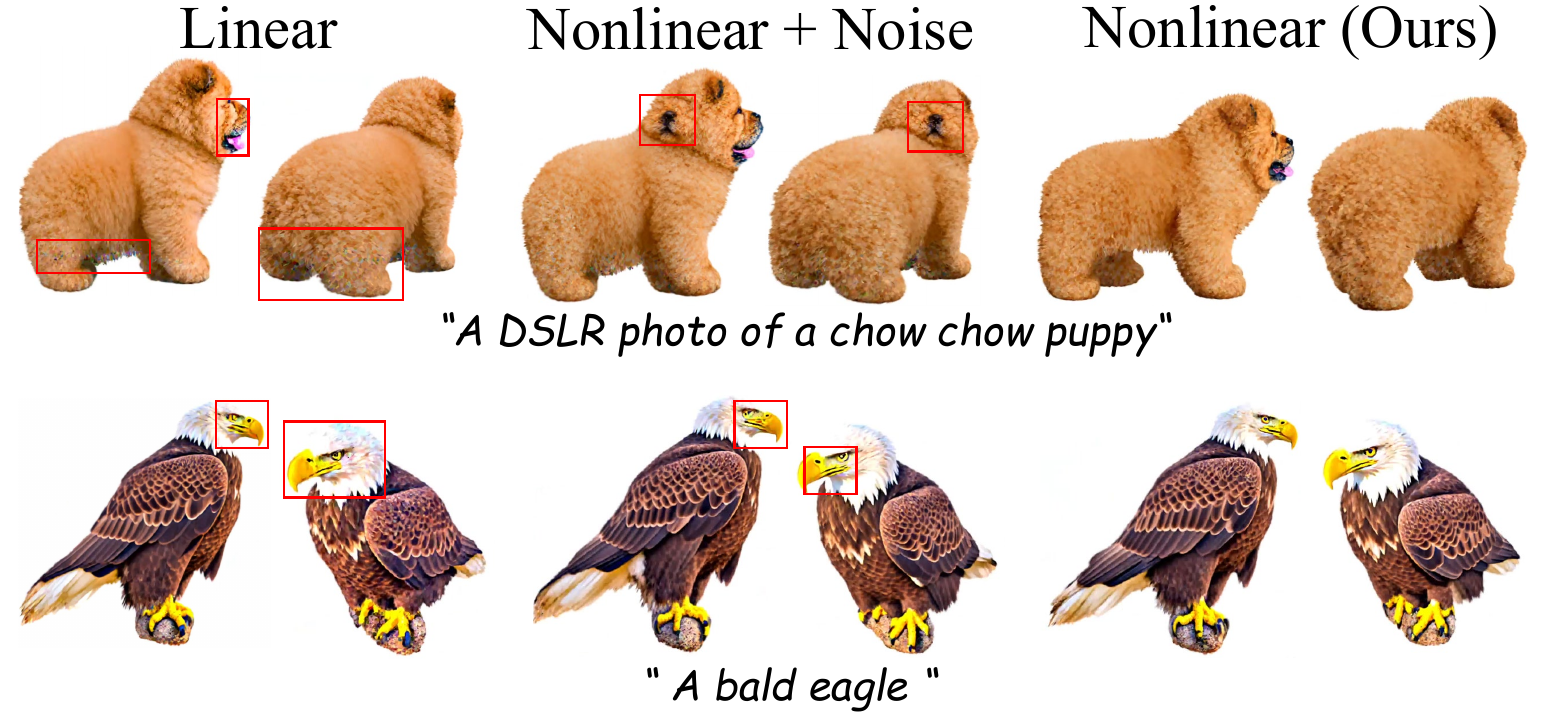}
    \caption{Ablation study on designs of the image degradation process. We show the effects of modeling this degradation with the linear learnable operator, nonlinear learnable operator with noise, and our choice, a noise-free nonlinear learnable operator ($M_\psi$).}
    \label{fig:ablation_2}
\end{figure}

\noindent\textbf{Discussion on the degradation process.} In Sec.~\ref{sec:vdm}, we model the degradation relationship between diffusion model-generated images and rendered images using a learnable nonlinear degradation operator without observation noise, which is the most important design of VDM. We evaluated three different degradation settings to validate our selection. Figure \ref{fig:ablation_2} reveals that nonlinear operators outperform linear ones, which are implemented as a learnable tensor matching the dimensions of $\epsilon_{\phi}(\mathbf{x}_{t}, t, y)$. 3D results from linear operators exhibit more noise due to limited modeling capacity.  Furthermore, the inclusion of learnable observation noise, denoted by a tensor equivalent in size to the predicted noise, fails to enhance 3D generation and may introduce artifacts in some instances.

\begin{figure}[!htb]
    \centering
    \includegraphics[width=1.0\linewidth]{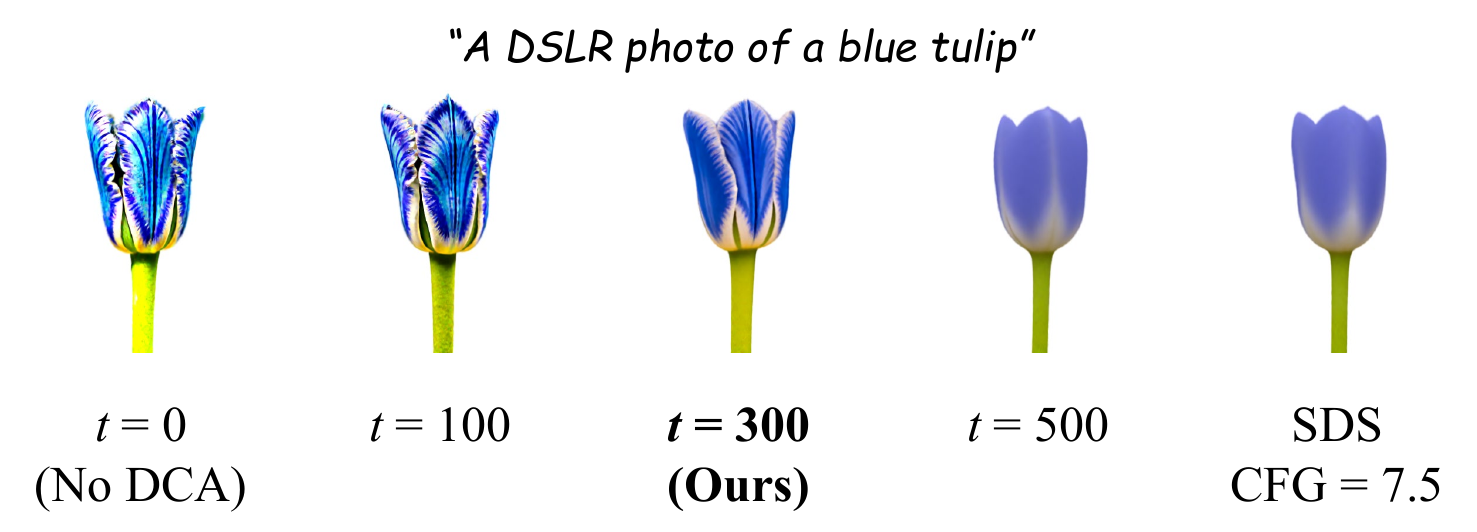}
    \caption{Ablation study on different timestep choice of DCA.}
    \label{fg:ablation DCA_t}
\end{figure}

\noindent\textbf{Choice of timestep in DCA.} In the execution of DCA, as outlined in Eq. (\ref{eq:DCA}), we select a specific timestep $t=300$ as the cut-off point to tune the coefficient $\lambda_{t}$ of rendered image distribution. As illustrated in Figure \ref{fg:ablation DCA_t}, we conduct experiments to substantiate the efficacy of our selection. When the timestep is set to 100, the generated 3D results are analogous to those without DCA, exhibiting color-saturation problems and some noise artifacts. Conversely, increasing the cut-off timestep to 500 leads to excess smoothness issues, and the rendering is similar to the result through the utilization of SDS with a CFG scaling factor of 7.5. Consequently, opting for a time step $t=300$ enables DreamMapping to generate high-quality 3D objects with realistic color and fewer artifacts.

\begin{figure}[!htb]
    \centering
    \includegraphics[width=1.0\linewidth]{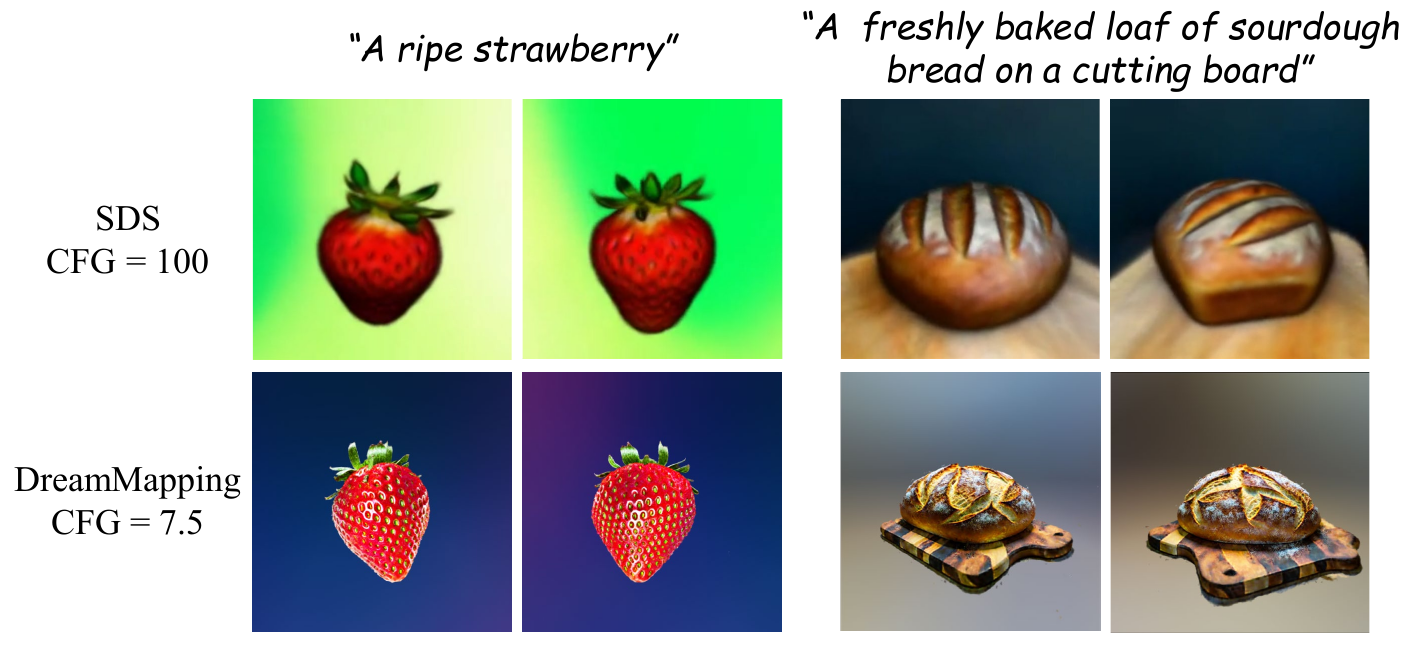}
    \caption{Generalizability of DreamMapping on NeRF.}
    \label{fig:ablation NeRF}
\end{figure}

\begin{figure}[!htb]
    \centering
    \includegraphics[width=1.0\linewidth]{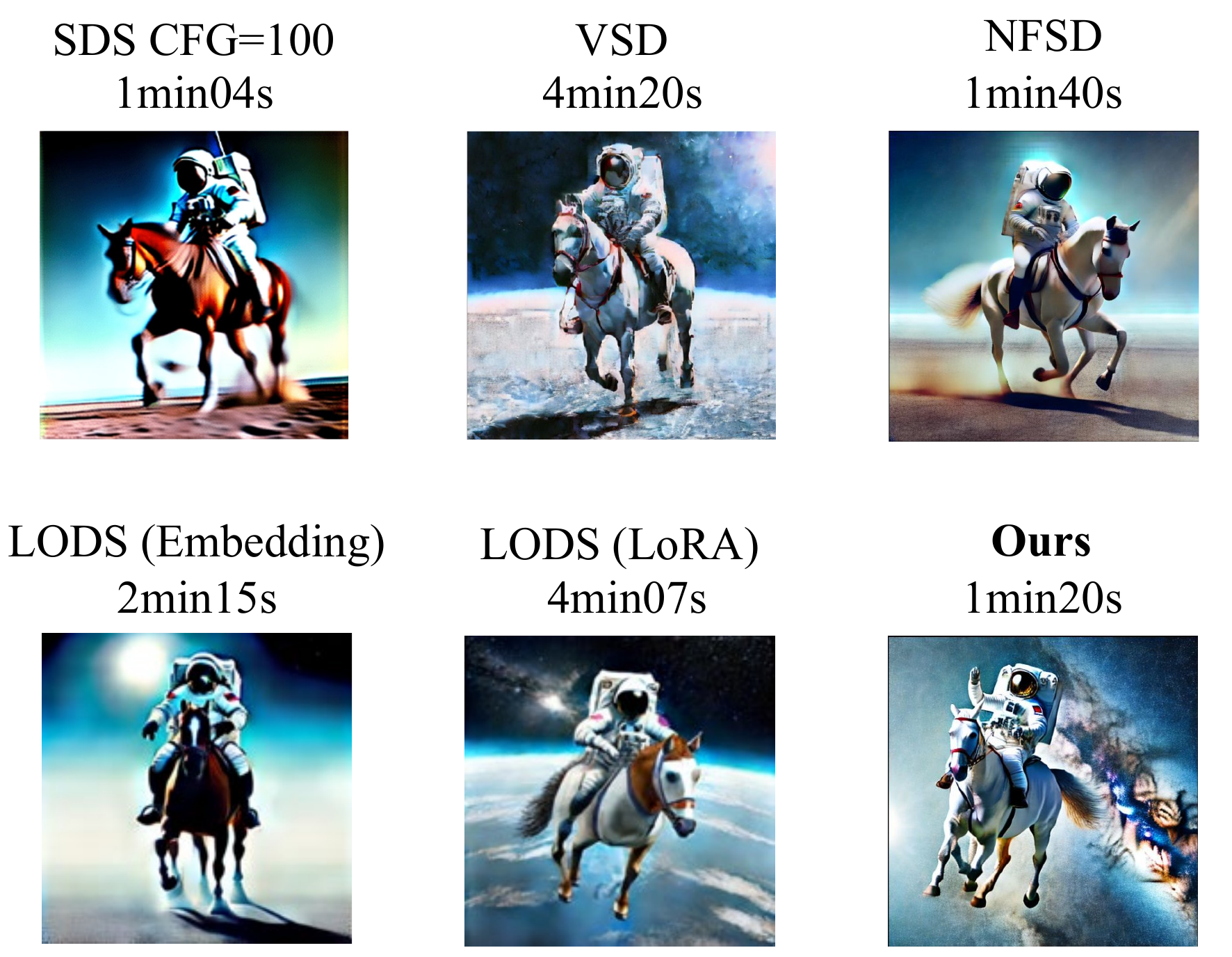}
    \caption{Qualitative comparisons for text-to-2D task. Our method displays clear details and mitigates color saturation issues. The experiment of generation time comparison was conducted on a single NVIDIA 4090 GPU.}  %less over saturation and no blurred appearance
    \label{fig:2D_Compare}
\end{figure}

%discuss the influence of 3D representation and 2D generation here.
\noindent \textbf{Generalizability of DreamMapping.} 
Although we integrate VDM and DCA into the 3DGS-based text-to-3D framework, their generative potential can be extended beyond this context. Our improvements on SDS can be applied to other 3D representations as well, e.g., NeRF~\cite{NeRF:ECCV:2020}. As depicted in Figure~\ref{fig:ablation NeRF}, we follow the hyperparameter configurations employed by ProlificDreamer~\cite{ProlificDreamer:NIPS:2023} in NeRF comparison experiments.Remarkably, our proposed method can deliver intricate details even under a CFG setting of 7.5. A particularly noteworthy aspect is the generation time afforded by our approach when integrated with NeRF, averaging approximately 1.9 hours, significantly shorter than ProfliciDreamer. This efficiency gain primarily stems from the construction of a variational distribution within our optimization framework, which bypasses the gradient computation for the Unet Jacobian term in diffusion models. Additionally, VDM and DCA methods can be seamlessly integrated into text-to-2D generation, exhibiting style-matching results with rich details and optimization efficiency, as shown in Figures~\ref{fig:2D_Compare} and~\ref{fig:add_2d_res}.

\section{Conclusion}
In this work, we have presented a comprehensive analysis and review of SDS-based methods. We further concluded their defects in dealing with the variational distribution. Based on the analysis, we have proposed Variational Distribution Mapping (VDM) with Distribution Coefficient Annealing (DCA), a novel approach for constructing an efficient variational distribution through a learnable neural network without refining the diffusion model. 

Building upon this, we have developed DreamMapping, a text-to-3D framework that integrates VDM and DCA with 3D Gaussian Splatting. Through extensive experiments and evaluations, we validate the effectiveness of our approach. Notably, our approach's compatibility can be extended to other 3D representations, e.g., NeRF and text-to-2D generation.

While DreamMapping can produce diverse high-fidelity 3D assets, it still has several limitations for further improvements. First, the generation quality of our framework relies heavily on geometry initialization. Second, the timestep choice of DCA could become a learnable factor, adapting to the diffusion model's dynamic characteristics. Besides, exploring the training of an independent image distribution model before the generation task may offer opportunities to expedite generation time in future research.

% discuss with other 3D representation
% discussed generalization (like 2D here)
%To improve ... distribution?, we intorduce VDM
%timestep

\section*{Acknowledgments}
We thank Dr. Ailing Zeng for the insightful discussions and all participants for evaluating our results. This project is partially supported by the CCF-Tencent Rhino-Bird Open Research Fund RAGR20230120 and the Open Project Program of the State Key Laboratory of CAD\&CG (Grant No. A2427), Zhejiang University.
\bibliographystyle{eg-alpha-doi} 
\bibliography{references}  
\clearpage
\section{Appendix} %supplemental materials part
%SSD - 
\label{sec:appendix}
\subsection{Analysis for DCA}
\label{sec: analysis for DCA}
Recalling our DCA discussion, we made a simplified assumption, using a timestep $t=300$ as the cut-off point for adjusting $\lambda$. Besides the ablation study in Sec.~\ref{sec:ablation}, here we substantiate this design and statement that the linear correlation between $\epsilon$ and $\epsilon_{\phi}(\textbf{x}_{t},t,y)$ becomes independent at a small timestep $t$.

As outlined in Eq. (\ref{Score Relationship}), the score function $\nabla_{\mathbf{x}_{t}}\log p(\mathbf{x}_{t}|y)$ in each timestep is constantly associated with $\epsilon_{\phi}(\textbf{x}_{t},t,y)$. Hence, as Tang et al.~\cite{SSD:Arxiv:2023} have validated, by demonstrating the linear correlation between $\epsilon$ and $\nabla_{\mathbf{x}_{t}}\log p(\mathbf{x}_{t}|y)$, we can indirectly prove the timestep-dependent linear relationship between $\epsilon$ and $\epsilon_{\phi}(\textbf{x}_{t},t,y)$. 

When $t=0$, we prove that $\nabla_{\mathbf{x}_{0}}\log p(\mathbf{x}_{0}|y)$ is independent of $\epsilon$. Since $x_0 = x + 0 \cdot \epsilon$, it is obvious that the inputs to $\nabla_{\mathbf{x}_{0}}\log p(\mathbf{x}_{0}|y)$ do not include any information about $\epsilon$. Therefore, $\epsilon_{\phi}(\textbf{x}_{0},t,y)$ is purely irrelated to $\epsilon$. When $t$ increases to the upper limit $t_{max}$, we prove that $\nabla_{\mathbf{x}_{t_{max}}}\log p(\mathbf{x}_{t_{max}}|y)$ is collinear to $\epsilon$ as follows:

\begin{align} 
    \begin{split}
    \nabla_{\mathbf{x}_{t_{max}}}\log p(\mathbf{x}_{t_{max}}|y) &=  \nabla_{x_{tmax}}log\mathcal{N}(x_{tmax};0, \mathcal{I}) \\ 
             &=  \nabla_{x_{tmax}}(-\frac{1}{2}x^{T}_{tmax}x_{tmax}) \\
             &= -\frac{1}{2}\nabla_{x_{tmax}}x^{T}_{tmax}x_{tmax} \\
             & = -x_{tmax} \\
             & = -0 \cdot x -1 \cdot \epsilon \\
             & = -\epsilon, 
    \label{eq:collinear}
    \end{split}
\end{align}
where the superscript $T$ represents the vector transpose operation. 

\subsection{More Visual Results}
\begin{figure*}[htb!]
    \centering    \includegraphics[width=0.9\linewidth]{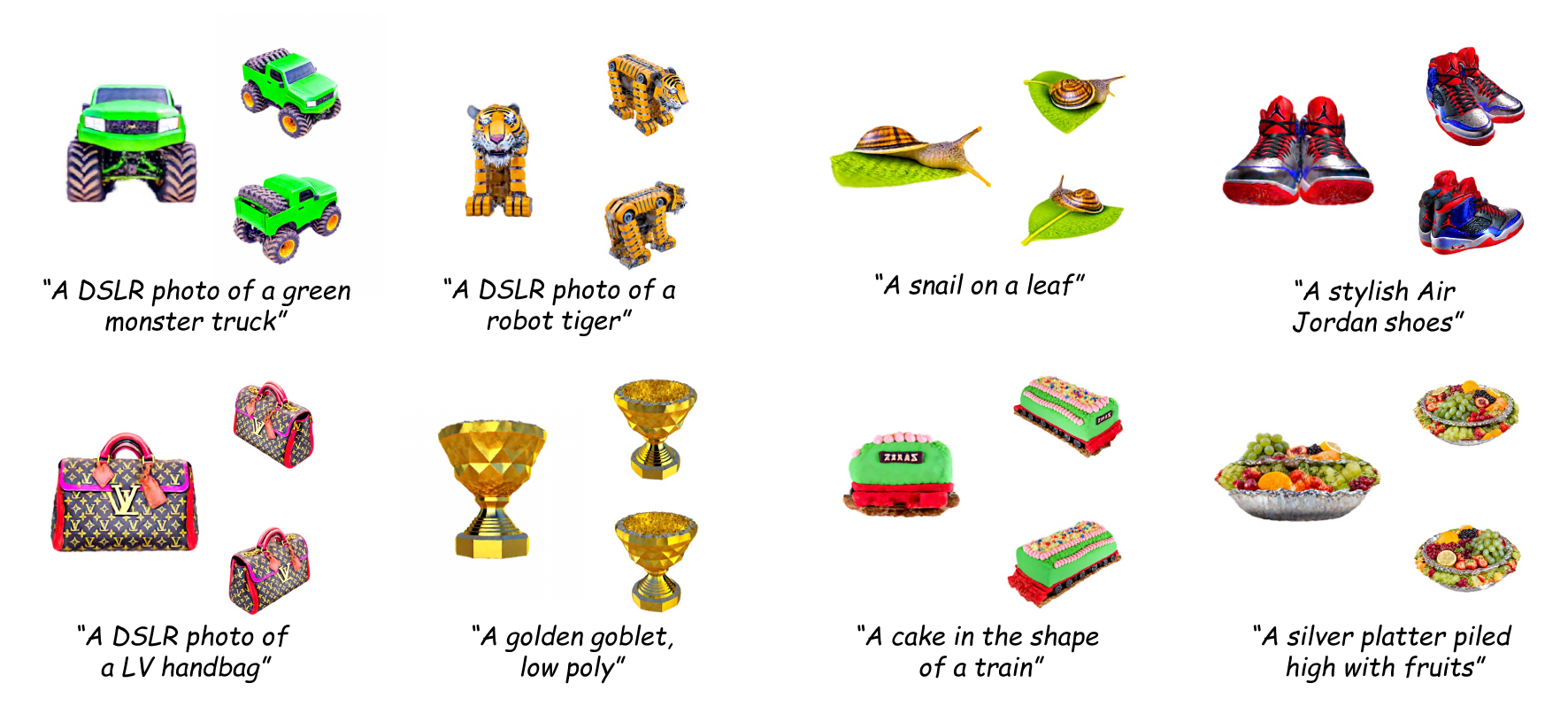}
    \caption{More text-to-3D results by our DreamMapping framework. Please zoom in for details.}
    \label{fig:add_res}
\end{figure*}

\begin{figure*}[!htb]
    \centering
   \includegraphics[width=0.95\linewidth]{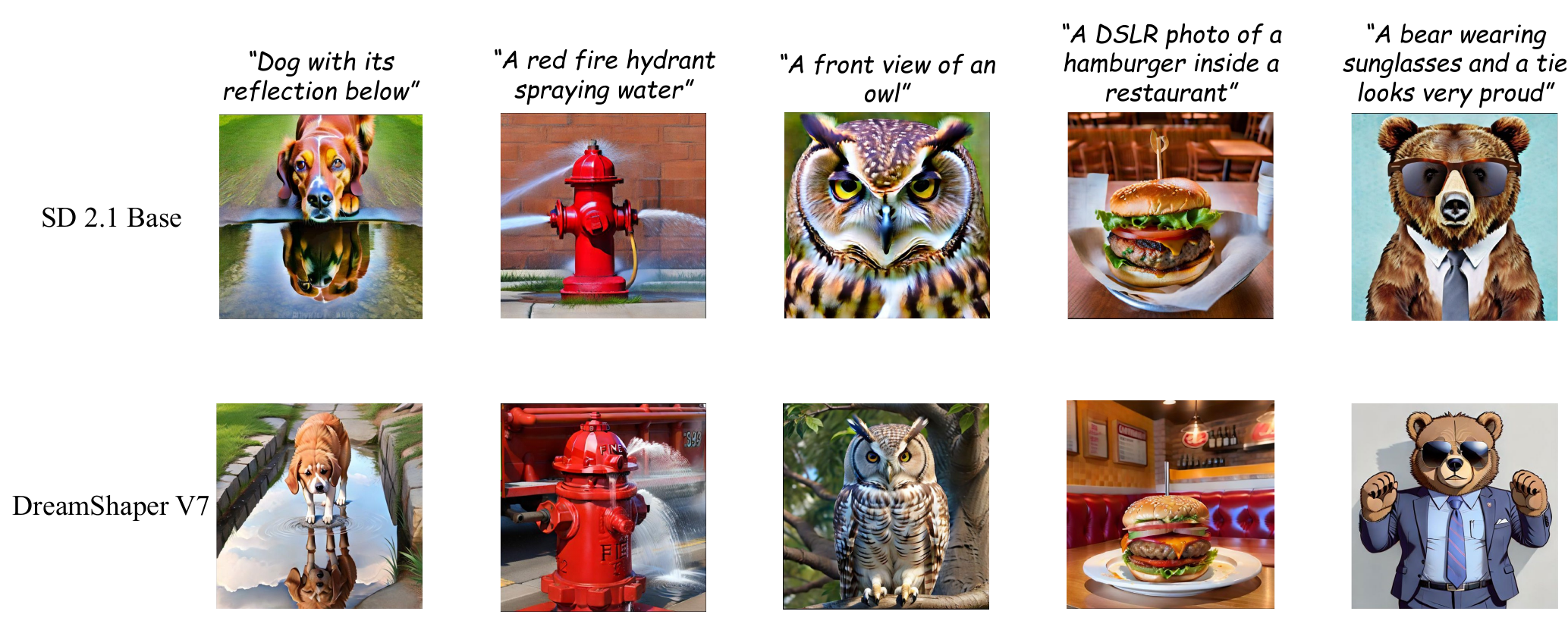}
    \caption{More text-to-2D results using different base models, i.e., SD 2.1 Base~\cite{SD-2-1-Base} and DreamShaper V7~\cite{DreamShaper-V7}.}
    \label{fig:add_2d_res}
\end{figure*}

We provide additional visual results in Figures~\ref{fig:add_res} and \ref{fig:add_2d_res}, illustrating DreamMapping's capacity in text-guided generation tasks.  

\begin{figure*}[!htb]
    \centering    
    \includegraphics[width=1.0\linewidth]{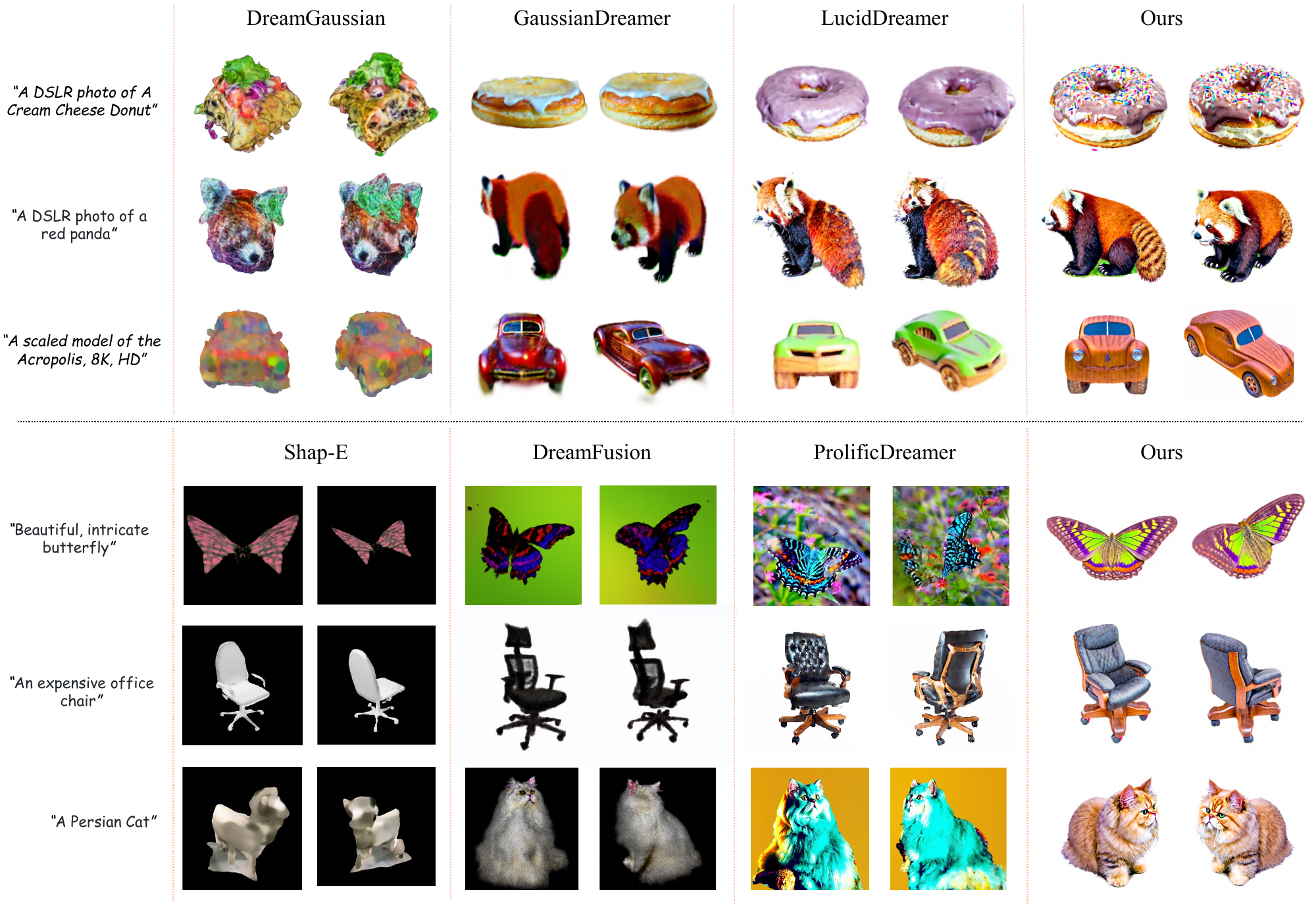}
    \caption{More qualitative comparisons with recent popular methods in text-to-3D generation. Please zoom in for details.}
    \label{fig:add_vis_comp}
\end{figure*}

\end{document}